\documentclass{article}

\usepackage{microtype}
\usepackage{graphicx}
\usepackage{subfigure}
\usepackage{booktabs} % for professional tables

\usepackage{hyperref}

\usepackage[accepted]{icml2023}

\usepackage{microtype}
\usepackage{subfigure}
\usepackage{booktabs} % for professional tables

\usepackage{dsfont}
\usepackage{amsmath, amssymb, amsfonts, amsthm}
\usepackage{mathtools,commath}
\usepackage{graphicx, multirow}
\usepackage{xcolor}
\usepackage{authblk}
\usepackage{hyperref}

\usepackage[capitalize,noabbrev]{cleveref}

\theoremstyle{plain}
\newtheorem{theorem}{Theorem}[section]
\newtheorem{proposition}[theorem]{Proposition}

\theoremstyle{definition}
\newtheorem{definition}[theorem]{Definition}

\theoremstyle{remark}

\usepackage[textsize=tiny]{todonotes}

\icmltitlerunning{Data-OOB: Out-of-bag Estimate as a Simple and Efficient Data Value}

\begin{document}

\twocolumn[
\icmltitle{Data-OOB: Out-of-bag Estimate as a Simple and Efficient Data Value}



\icmlsetsymbol{equal}{*}

\begin{icmlauthorlist}
\icmlauthor{Yongchan Kwon}{columbia}
\icmlauthor{James Zou}{stanford,aws}
\end{icmlauthorlist}

\icmlaffiliation{columbia}{Columbia University}
\icmlaffiliation{stanford}{Stanford University}
\icmlaffiliation{aws}{Amazon AWS}

\icmlcorrespondingauthor{James Zou}{jamesz@stanford.edu}

\icmlkeywords{Data valuation, Random Forest, Bagging, Data-centric Machine Learning}

\vskip 0.3in
]



\printAffiliationsAndNotice{\icmlEqualContribution} % otherwise use the standard text.

\begin{abstract}
Data valuation is a powerful framework for providing statistical insights into which data are beneficial or detrimental to model training. Many Shapley-based data valuation methods have shown promising results in various downstream tasks, however, they are well known to be computationally challenging as it requires training a large number of models. As a result, it has been recognized as infeasible to apply to large datasets. To address this issue, we propose Data-OOB, a new data valuation method for a bagging model that utilizes the out-of-bag estimate. The proposed method is computationally efficient and can scale to millions of data by reusing trained weak learners. Specifically, Data-OOB takes less than $2.25$ hours on a single CPU processor when there are $10^6$ samples to evaluate and the input dimension is $100$. Furthermore, Data-OOB has solid theoretical interpretations in that it identifies the same important data point as the infinitesimal jackknife influence function when two different points are compared. We conduct comprehensive experiments using 12 classification datasets, each with thousands of sample sizes. We demonstrate that the proposed method significantly outperforms existing state-of-the-art data valuation methods in identifying mislabeled data and finding a set of helpful (or harmful) data points, highlighting the potential for applying data values in real-world applications.
\end{abstract}

\section{Introduction}
\label{sec:introduction}
% Motivation
Assessing the impact of data on a model's performance is important as it enhances our understanding of the data. Moreover, it has various practical real-world applications including medical image analysis, data curation, and data marketplaces \citep{tang2021data, agarwal2019, tian2022private}. Due to its importance, data valuation has become a primary research topic in machine learning and statistics. The main goal is to establish a practical and principled notion of the influence of individual data points on the process of training a model.

% Introduction of existing methods
A standard approach for evaluating the impact of data is to use the marginal contribution, which is defined as the average change in a model's performance when a certain datum is removed from a set of data points. One approach uses the leave-one-out (LOO) method, which evaluates a single marginal contribution when a certain datum is removed from the entire training dataset \citep{cook1982residuals, koh2017understanding}. Another approach is based on the Shapley value in cooperative game theory \citep{shapley1953}, assigning a simple average of all marginal contributions by varying the number of data points in a given subset of the training dataset. Data Shapley \citep{ghorbani2019}, Distributional Shapley \citep{ghorbani2020distributional}, and CS-Shapley \citep{schoch2022cs} belong to this category.

% Critical problems of existing Shapley-based methods
Existing works have shown that Shapley-based methods perform better than LOO in many downstream tasks by leveraging every possible marginal contribution \citep{ghorbani2019, jia2019b}. However, it often requires training a significant number of models to accurately estimate marginal contributions. This has been recognized as the primary limitation in practical applications of data valuation.
To address this issue, improved approximation algorithms using stratified sampling have been developed \citep{maleki2013bounding, wu2022robust}, but they still require training multiple models to achieve a small enough approximation error. As an alternative approach, data values with a closed-form expression have been studied. It can easily scale up to millions of data but restricts the model to be either $k$-Nearest Neighbors \citep{jia2019b} or linear models \citep{kwon2021efficient}, which may not be most suitable for high dimensional data analysis. Recently, \citet{lin2022measuring} proposed an algorithm that uses the linear coefficient of a LASSO model. Their method is shown to have improved sample efficiency over existing methods, but it still requires unverifiable assumptions and additional computational costs for training LASSO models.

In addition to the computational challenges, the Shapley value and its variants have critical limitations. They are based on the fair division axioms in cooperative game theory, but the axioms' relevance to machine learning applications is unclear \citep{sim2022data, rozemberczki2022shapley}. 

\begin{figure}
    \centering
    \includegraphics[width=1.025\columnwidth]{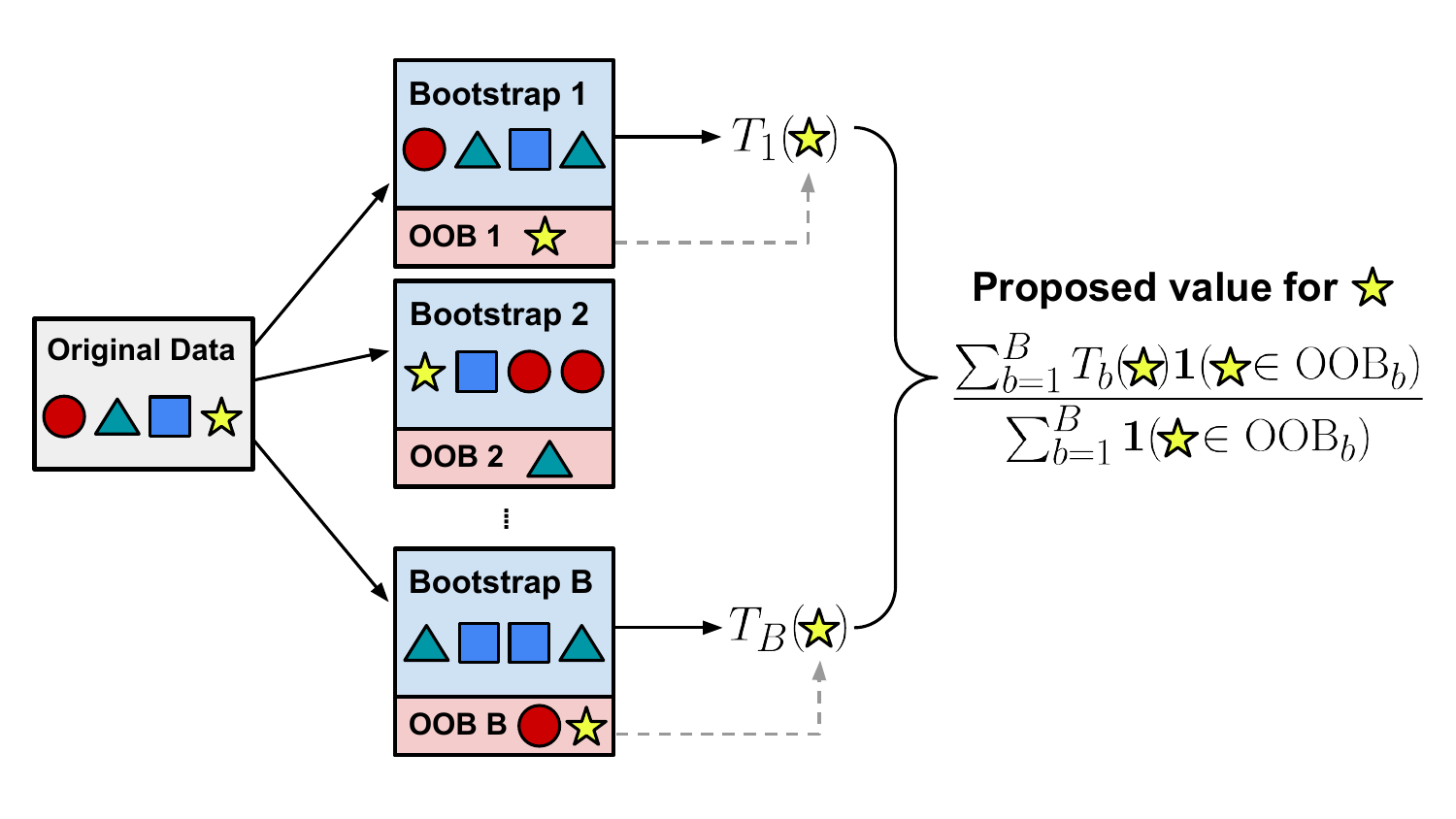}
    \vspace{-0.25in}
    \caption{Illustration of the proposed data valuation method. The OOB stands for the out-of-bag set. For each bootstrap sampling procedure, we evaluate an estimate $T_b(\star)$ if the datum $\star$ is in the OOB set. Here, $T_b(\star)$ is a score of the model trained with the $b$-th bootstrap dataset evaluated at $\star$. The proposed data value summarizes scores $T_b(\star)$ from the $B$ bootstrap datasets. Details are provided in Section~\ref{sec:proposed}.} 
    \label{fig:illustration_of_method}
    \vspace{-0.1in}
\end{figure}

% Proposed
\paragraph{Our contributions} In this paper, we propose Data-OOB, a new data valuation framework for a bagging model that uses the out-of-bag (OOB) estimate as illustrated in Figure~\ref{fig:illustration_of_method}. Our framework is computationally efficient by leveraging trained weak learners and is even faster than KNN-Shapley which has a closed-form expression. Furthermore, Data-OOB is statistically interpretable in that under mild assumptions it identifies the same important data point as the infinitesimal jackknife influence function when two different points are compared. Our comprehensive experiments demonstrate that the proposed method significantly better identifies mislabeled data and determines which data points are beneficial or detrimental for a model's performance than existing state-of-the-art data valuation methods. 

\section{Preliminaries}
\label{sec:preliminary}
For $d \in \mathbb{N}$, we denote an input space and an output space by $\mathcal{X} \subseteq \mathbb{R}^{d}$ and $\mathcal{Y} \subseteq \mathbb{R}$, respectively. We denote a training dataset by $\mathcal{D}=\{(x_i, y_i)\}_{i=1} ^n$ where $x_i \in \mathcal{X}$ and $y_i \in \mathcal{Y}$ are an input and its label for the $i$-th datum. We denote a utility function by $U$, which takes as input a subset of the training dataset $\mathcal{D}$ and outputs  a model's performance that is trained on that subset. In classification, for instance, a common choice for $U$ is the test classification accuracy of an empirical risk minimizer trained on a subset of $\mathcal{D}$, \textit{i.e.}, $U(S) = \mathbb{E}[ \mathds{1}(Y = \hat{f}_S(X))]$ where $\mathds{1}(A)$ is an indicator whose value is one if a statement $A$ is true and zero otherwise, $\hat{f}_S := \mathrm{argmin}_{f \in \mathcal{F}} \sum_{j \in S} \mathds{1}( y_j \neq f(x_j))$ for some class of models $\mathcal{F}=\{f | f : \mathcal{X} \to \mathcal{Y} \}$, and the expectation $\mathbb{E}$ is taken with respect to a data distribution or is often approximated by a finite holdout validation dataset. When $S=\{\}$ is the empty set, $U(S)$ is set to be the performance of the best constant model by convention. 
A utility function depends on the choice of learning algorithms and a class $\mathcal{F}$, but we suppress its dependency as our main focus is on comparing the functional form of data values. For a set $S$, we denote its power set by $2^{S}$ and its cardinality by $|S|$. We set $[j]:=\{1, \dots, j\}$ for $j \in \mathbb{N}$. 

A standard approach for quantifying data values is to use the marginal contribution, which measures the average change in a utility function when a particular datum is removed from a subset of the entire training dataset $\mathcal{D}$. 
\begin{definition}[Marginal contribution]
For a utility function $U:2^{\mathcal{D}} \to \mathbb{R}$ and $j \in [n]$, the marginal contribution of $z \in \mathcal{D}$ with respect to $j$ samples is defined as follows.
\begin{align*}
    \Delta_j (z, U) := \frac{1}{\binom{n-1}{j-1}} \sum_{ S \in \mathcal{D}_{j} ^{\backslash z} } U(S\cup \{z\})-U(S),
\end{align*}
where $\mathcal{D}_{j} ^{\backslash z} := \{ S \subseteq \mathcal{D} \backslash \{z\}: |S|=j-1 \}$.
\label{def:marginal_contrib}
\end{definition}
Many data valuation methods can be expressed as a function of the marginal contribution. The LOO method is $\Delta_n (z, U)$, measuring the changes when one particular datum $z$ is removed from the entire dataset $\mathcal{D}$. LOO includes the Cook's distance and the approximate empirical influence function \citep{cook1980characterizations, koh2017understanding}. Another example is Data Shapley \citep{ghorbani2019}, which is expressed as a simple average of marginal contributions $\psi_{\mathrm{Shap}}(z, U) := n^{-1} \sum_{j=1} ^n \Delta_j (z, U)$. As its extension, Beta Shapley proposed by \citet{kwon2022beta} is expressed as a weighted mean of marginal contributions.
\begin{align}
    \psi_{\mathrm{Beta}}(z, U, \beta) := \sum_{j=1} ^n \beta_j \Delta_j (z, U),
    \label{eqn:beta_shap}
\end{align}
where $\beta=(\beta_1, \dots, \beta_n)$ is a predefined weight vector such that $\sum_{j=1} ^n \beta_j=1$ and $\beta_j \geq 0$ for all $j \in [n]$. A functional form of Equation~\eqref{eqn:beta_shap} is also known as semivalues in cooperative game theory. 

The LOO method is known to be computationally feasible, but it often assigns erroneous values that are close to zero \citep{basu2020influence}. Data Shapley and Beta Shapley are empirically shown to be more effective than LOO in many downstream tasks such as mislabeled data detection in classification settings \citep{ghorbani2019,kwon2022beta}. However, their computational complexity is well known to be expensive, making it infeasible to apply to large datasets  \citep{bachrach2010approximating, jia2019, wang2022data}. As a result, most existing work has focused on small datasets, \textit{e.g.}, $n \leq 1000$. 

Many methods have been proposed to reduce computational costs. \citet{wu2022robust} proposed a stratified sampling to optimize the number of utility evaluations, and \citet{jia2019b} and \citet{kwon2021efficient} derived a closed-form expression of the Shapley value. However, these methods still have difficulties in scaling to large datasets or require unusual assumptions on the utility function. For instance, \citet{jia2019b} used a utility function that does not take into account the majority voting in classification settings: for $S \in \mathcal{D}$, $U(S) = k^{-1} \sum_{(x_i, y_i) \in \mathcal{N}(S)} \mathds{1}( \tilde{y} = y_i)$ where $\mathcal{N}(S)$ is a set of $\min(k, |S|)$ nearest neighbors of $\tilde{x}$, and $(\tilde{x}, \tilde{y}) \in \mathcal{X} \times \mathcal{Y}$ is a test datum. \citet{kwon2021efficient} considered a commonly used utility function (\textit{e.g.}, the negative Euclidean distance), but it is limited to linear regression models, which may not be the most favorable in real-world data analysis.

Recently, \citet{lin2022measuring} proposed an efficient algorithm to estimate a class of data values called the average marginal effect (AME) given as follows.
\begin{align*}
    \psi_{\mathrm{AME}} (z, U) := \mathbb{E}_{S}[U(S\cup\{z\})-U(S)],
\end{align*}
where the expectation $\mathbb{E}_{S}$ is taken over a random set $S$ with a user-defined distribution defined on the discrete space $\cup_{j=1} ^n \mathcal{D}_{j} ^{\backslash z}$. They showed that AME can include the Shapley value and semivalues as a special case, and it can be approximated by the linear coefficient of a LASSO model. Specifically, AME is estimated by a minimizer of the following objective function.
\begin{align*}
    \mathrm{argmin}_{\gamma \in \mathbb{R}^n} \frac{1}{|\mathcal{S}|} \sum_{S\in \mathcal{S}} \left( U(S) - g( \mathds{1}_{S} )^T \gamma \right)^2 + \lambda \sum_{i=1} ^n |\gamma_i|,
\end{align*}
where $\lambda > 0$ is a regularization parameter, $g: \{0,1\}^n \to \mathbb{R}^n$ is a predefined transformation function, $\mathcal{S} = \{S : S \subseteq \mathcal{D} \}$ is a set of subsets of $\mathcal{D}$ and $\mathds{1}_S \in \{0,1\}^n$ is $n$-dimensional vector whose element is one if its index is an element of $S$, zero otherwise. 
Their algorithm is shown to have better computational efficiency in the semivalue estimation than existing methods. However, it requires some sparsity assumption that is difficult to verify, and also, it needs training a LASSO model. 

\citet{ilyas2022datamodels} proposed a similar idea called datamodels that use a LASSO model to predict a test data point's prediction, \textit{i.e.}, the utility $U(S)$ is evaluated at a particular test data point. Due to its dependency on a particular datum, it is not suitable for capturing the influence on the model performance. Moreover, it needs computational costs for training a LASSO model similar to \citet{lin2022measuring}.

Besides the computational issue, marginal contribution-based methods have another critical issue with theoretical interpretation. Motivated by cooperative game theory, they provide mathematical justifications that are seemingly solid. However, the fair division axioms used in the Shapley value have not been statistically examined, and it raises a fundamental question about the appropriateness of these axioms in machine learning problems \citep{kumar2020problems, sim2022data, rozemberczki2022shapley}. 

In the following section, we propose a novel data valuation framework for a bagging model that can address aforementioned issues. Our method is computationally efficient by recycling trained weak learners and does not rely on the fair division axioms that can be less relevant to machine learning applications.

\section{Data-OOB: Out-Of-Bag Estimate as Data Value}
\label{sec:proposed}
Suppose we have a trained bagging model that consists of $B$ weak learner models. For $b \in [B]$, we denote the $b$-th weak learner by $\hat{f}_b:\mathcal{X} \to \mathcal{Y}$, which is trained on the $b$-th bootstrap dataset. It can be expressed as a minimizer of a weighted risk as follows.
\begin{align*}
    \hat{f}_b := \mathrm{argmin}_{f \in \mathcal{F}} \frac{1}{n} \sum_{j=1} ^n w_{bj} \ell(y_{j} , f(x_{j})),
\end{align*}
where $\ell:\mathcal{Y} \times \mathcal{Y} \to \mathbb{R}$ is a loss function and $w_{bj} \in \mathbb{Z}$ is the number of times the $j$-th datum $(x_j, y_j)$ is selected in the $b$-th bootstrap dataset. We set $w_b:=(w_{b1}, \dots, w_{bn})$ for all $b \in [B]$. For $i\in[n]$ and $\Theta_B :=\{(w_b, \hat{f}_b)\}_{b=1} ^B$, we propose to use the following quantity as data values. 
\begin{align}
    \psi ((x_i, y_i), \Theta_B) := \frac{\sum_{b=1} ^B \mathds{1}(w_{bi} =0) T(y_i, \hat{f}_b (x_i)) }{\sum_{b=1} ^B \mathds{1}(w_{bi} =0)}.
    \label{eqn:proposed_oob}
\end{align}
where $T: \mathcal{Y} \times \mathcal{Y} \to \mathbb{R}$ is a score function that represents the goodness of a weak learner $\hat{f}_b$ at the $i$-th datum $(x_i, y_i)$. For instance, we can use the correctness function $T(y_i, \hat{f}_b(x_i)) = \mathds{1}(y_i = \hat{f}_b (x_i))$ in classification settings and the negative Euclidean distance $T(y_i, \hat{f}_b (x_i)) = -(y_i - \hat{f}_b (x_i))^2$ in regression settings.

Our proposed data value in Equation~\eqref{eqn:proposed_oob} measures the average score when the datum $(x_i, y_i)$ is not selected in the bootstrap dataset. Accordingly, it can be interpreted as a partition of the OOB estimate, which is originally introduced to estimate the prediction error \citep{efron1992jackknife, efron1997improvements}. Specifically, the OOB estimate is given as
\begin{align*}
    \frac{1}{n} \sum_{i=1} ^n \frac{\sum_{b=1} ^B \mathds{1}(w_{bi} =0) T(y_i, \hat{f}_b (x_i)) }{\sum_{b=1} ^B \mathds{1}(w_{bi} =0)},
\end{align*}
and it is equal to the simple average of the proposed data values $\frac{1}{n} \sum_{i=1} ^n \psi ((x_i, y_i), \Theta_B)$. Motivated by this relationship, we call our data valuation method Data-OOB.

Data-OOB has several advantages in computational efficiency. In contrast to existing marginal contribution-based data valuation methods, Data-OOB can leverage trained weak learners $\hat{f}_b$. In other words, Data-OOB does not require training multiple models for the utility evaluation and is readily obtained when there is a trained bagging model. Moreover, it has sample efficiency because it does not use additional validation data points for the utility evaluation that can greatly affect the quality of data values. 

\paragraph{Theoretical interpretation}
We rigorously examine the statistical implications of our proposed method. We show that Data-OOB identifies the same set of important data points as Jaeckel's infinitesimal jackknife influence function \citep{jaeckel1972infinitesimal, mallows1975some}. 
We denote the empirical distribution of $\mathcal{D}$ by $\hat{\mathbb{P}} = \frac{1}{n}\sum_{j=1} ^n \delta_{(x_j, y_j)}$ where $\delta_{(x, y)}$ is the Dirac delta measure on $(x,y) \in \mathcal{X} \times \mathcal{Y}$. We reformulate the OOB estimate as a functional defined on a probability measure space: for a probability measure $\mathbb{Q}$ defined on $\mathcal{D}$, 
\begin{align*}
    h(\mathbb{Q}):= \int \psi ((x, y), \Theta_B) d\mathbb{Q}(x,y).
\end{align*}
Then, the infinitesimal jackknife influence function of $h$ is defined as its derivative. For $-1/(n-1) < \varepsilon < 1$, 
\begin{align*}
    \psi_{\mathrm{IJ}}(x_i,y_i) := \left.\frac{\partial h(\hat{\mathbb{P}}_{\varepsilon, i}) }{\partial \varepsilon} \right|_{\varepsilon=0} = \lim_{\varepsilon \to 0} \frac{ h(\hat{\mathbb{P}}_{\varepsilon, i}) - h(\hat{\mathbb{P}})}{\varepsilon},
\end{align*}
where $\hat{\mathbb{P}}_{\varepsilon, i} = (1-\varepsilon) \hat{\mathbb{P}} + \varepsilon \delta_{(x_i, y_i)}$. The infinitesimal jackknife influence function $\psi_{\mathrm{IJ}}(x_i,y_i)$ quantifies how fast the OOB estimate $h$ changes when the weight on the $i$-th datum $(x_i,y_i)$ is changed. Given that the OOB estimate $h$ approximates the test performance, $\psi_{\mathrm{IJ}}(x_i,y_i)$ is expected to capture the influence of individual data points on the test performance.

Although the name suggests similarity, it is important to note that the $\psi_{\mathrm{IJ}}$ is distinct from the empirical influence function widely studied in the machine learning literature \citep{koh2017understanding, basu2020influence, feldman2020neural}. Specifically, $\psi_{\mathrm{IJ}}$ measures how fast the OOB estimate changes, but the ordinary influence function measures how fast the test accuracy evaluated on several test data points changes. Moreover, when it comes to the functional form, $\psi_{\mathrm{IJ}}$ is defined without dependency on test data points, but the ordinary influence function requires them. We also highlight that the OOB estimate has not been studied much in the field of data valuation, even though its derivative $\psi_{\mathrm{IJ}}(x_i,y_i)$ intuitively describes the influence of individual data points on the model performance. 

The following proposition shows that the influence function and the proposed method identify the same important data point among two different data points under a mild condition. To begin with, we set $V_B := B^{-1} \sum_{b=1} ^B (q_b - \bar{q})^2$ where $q_b = \frac{1}{n}\sum_{j=1} ^n \mathds{1}(w_{bj}=0) T(y_j, \hat{f}_b (x_j))$ is the normalized OOB score for the $b$-th bootstrap dataset and $\bar{q} := B^{-1} \sum_{b=1} ^B q_b$.

\begin{proposition}[Order consistency between Data-OOB and the infinitesimal influence function]
For $i \neq j \in [n]$, if $\psi_{\mathrm{IJ}}(x_i,y_i) > \psi_{\mathrm{IJ}}(x_j,y_j) + 4 \sqrt{2} V_B ^{1/2}$, then $\psi ((x_{i}, y_{i}), \Theta_B) > \psi ((x_{j}, y_{j}), \Theta_B)$.
\label{prop:consistent_ordering}
\end{proposition}

A proof is given in Appendix~\ref{sec:proof}.
Proposition~\ref{prop:consistent_ordering} provides new statistical insights when two data points are compared: The proposed method and the infinitesimal jackknife influence function have the order if one data point has a large enough influence function value than the other. Here, $V_B$ is the variance of OOB scores $q_b$ across different bootstrap datasets, and it is expected to be very small (\textit{e.g.}, $O(n^{-1})$) when $n$ and $B$ are large enough \citep{efron1979bootstrap}. In short, when there is a large enough gap between two influence function values, the proposed method will have the same ordering. Given that many applications of data valuation mainly focus on the order of data points, this theoretical result highlights the potential efficacy of the method in downstream tasks.  

\section{Experiments}
\label{sec:experiment}
In this section, we systematically investigate the practical effectiveness of the proposed data valuation method \texttt{Data-OOB} through three sets of experiments, which are frequently used in previous studies: time comparison, mislabeled data detection, and point removal experiment. We demonstrate that our method is computationally efficient and highly effective in identifying mislabeled data. Furthermore, compared to state-of-the-art data valuation methods, \texttt{Data-OOB} better determines which data points are beneficial or detrimental for model training. 

\paragraph{Experimental settings}
We use 12 classification datasets that are publicly available in OpenML \citep{feurer-arxiv19a} or the Python package `scikit-learn' \citep{Pedregosa2011Scikit}, and have at least $15000$ samples. Also, we note that many of these datasets were used in previous data valuation papers \citep{ghorbani2019, kwon2022beta}. We compare \texttt{Data-OOB} with the following four data valuation methods: \texttt{KNN Shapley} \citep{jia2019b}, \texttt{Data Shapley} \citep{ghorbani2019}, \texttt{Beta Shapley} \citep{kwon2022beta}, and \texttt{AME} \citep{lin2022measuring}. We set the training sample size to $n \in \{1000, 10000\}$, but \texttt{Data Shapley} and \texttt{Beta Shapley} are computed only when $n=1000$ due to their low computational efficiency. All methods except for \texttt{Data-OOB} require additional validation data to evaluate the utility function $U$. We set the validation sample size to 10\% of the training sample size $n$. As for \texttt{Data-OOB}, we use a random forest model with $B=800$ decision trees. To make our comparison fair, we use the same number or a greater number of utility evaluations for \texttt{Data Shapley}, \texttt{Beta Shapley}, and \texttt{AME} compared to \texttt{Data-OOB}. Implementation details are provided in Appendix~\ref{app:implementation_details}. 

\subsection{Elapsed Time Comparison}
\label{sec:time_comparison}
We first assess the computational efficiency of \texttt{Data-OOB} using a synthetic binary classification dataset. For $d \in\{10, 100\}$, an input $X \in \mathbb{R}^d$ is randomly generated from a multivariate Gaussian distribution with zero mean and an identity covariance matrix, and an output $Y \in \{0,1\}$ is generated from a Bernoulli distribution with a success probability $p(X)$. Here, $p(X) := 1/(1+\exp(-X^T \eta))$ and each element of $\eta \in \mathbb{R}^d$ is generated from a standard Gaussian distribution. We only generate $\eta$ once, and the same $\eta$ is used to generate different data points. A set of sample sizes $n$ is $\{10^4, 2.5 \times 10^4, 5 \times 10^4, 10^5, 2.5 \times 10^5, 5 \times 10^5\}$. We measure the elapsed time with a single Intel Xeon E5-2640v4 CPU processor. For a fair comparison, the elapsed time for \texttt{Data-OOB} includes the training time for the random forest.

\begin{figure}[t]
    \centering
    \includegraphics[width=0.485\columnwidth]{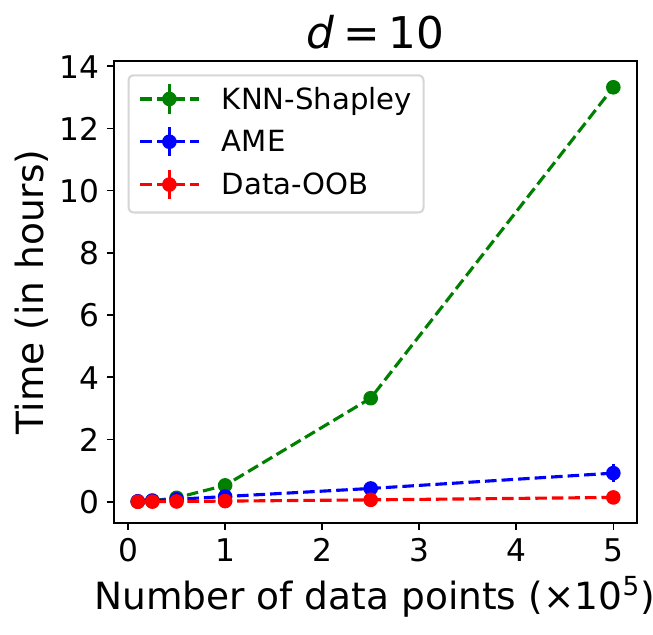}
    \includegraphics[width=0.495\columnwidth]{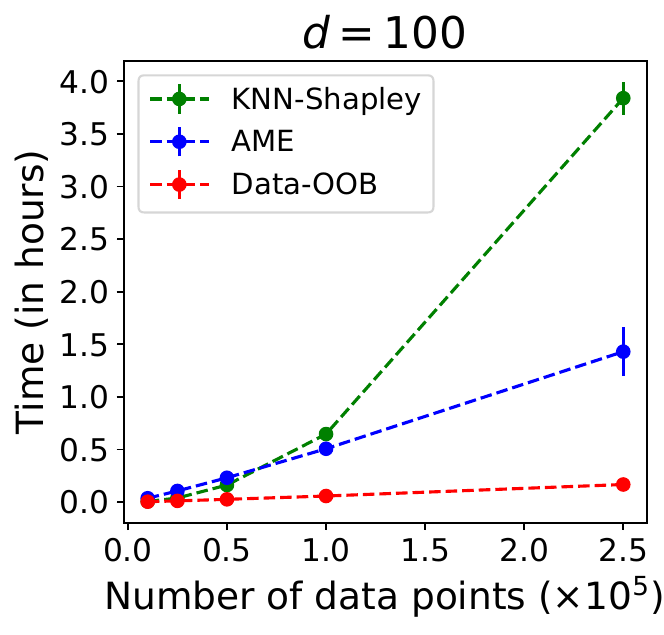}
    \vspace{-0.075in}
    \caption{Elapsed time comparison between \texttt{KNN Shapley}, \texttt{AME}, and \texttt{Data-OOB}. We use a synthetic binary classification dataset with (left) $d=10$ and (right) $d=100$. We exclude the setting $(n,d)=(5\times 10^5, 100)$ as \texttt{KNN Shapley} exceeds 24 hours. The error bar indicates a 95\% confidence interval based on 5 independent experiments. \texttt{Data-OOB} is significantly faster than \texttt{KNN Shapley} and \texttt{AME}. The time for training the random forest is included in the time for \texttt{Data-OOB}.}
    \vspace{-0.125in}
    \label{fig:time_comparison}
\end{figure}

As Figure~\ref{fig:time_comparison} shows, \texttt{Data-OOB} achieves better computational efficiency than existing methods \texttt{KNN Shapley} and \texttt{AME} in various $n$ and $d$. Specifically, \texttt{Data-OOB} is $54$ times faster than \texttt{KNN Shapley} when $(n,d)=(10^5, 10)$. Interestingly, we find \texttt{KNN Shapley} is slow despite having the closed-form expression because it needs to sort $n$ data points for each validation data point. When $(n,d)=(5 \times 10^5, 100)$ and the validation sample size is $10^4$, \texttt{KNN Shapley} exceeds 24 hours. For this reason, we exclude this setting from Figure~\ref{fig:time_comparison}. \texttt{KNN Shapley} can be more efficient if the validation size is smaller, but it would cost the quality of data values. In comparison with \texttt{AME}, \texttt{Data-OOB} does not require training LASSO models, achieving better computational efficiency. 

As for the algorithmic complexity, when a random forest is used, the computational complexity of Data-OOB will be $O(Bdn\log(n))$ where $B$ is the number of trees, $d$ is the number of features and $n$ is the number of data points in the training dataset. This is because the computational cost of \texttt{Data-OOB} is mainly from training a random forest model, and its computational complexity is $O(Bdn\log(n))$ \citep{hassine2019important}. Meanwhile, the computational complexity of \texttt{KNN Shapley} will be $O(n^2\log(n))$ when the number of data points in the validation dataset is $O(n)$ (e.g. 10\% of $n$). These results support why the elapsed time for \texttt{Data-OOB} increases linearly and that of the \texttt{KNN-Shapley} increases polynomially in Figure~\ref{fig:time_comparison}. In addition, it shows that ours can be beneficial when $n$ is increasing but $B$ and $d$ are fixed.

Our method is highly efficient and it takes less than 2.25 hours when $(n,d)=(10^6, 100)$ on a single CPU processor. The proposed method can be more efficient with the use of trained multiple weak learners. For instance, when $(n,d)=(10^5, 10)$, the computation of \texttt{Data-OOB} takes only 13\% of the entire training time for a random forest. 

\subsection{Mislabeled Data Detection}
\label{sec:mislabeled_data_experiment}

\begin{figure*}[t]
    \centering
    \includegraphics[width=0.24\textwidth]{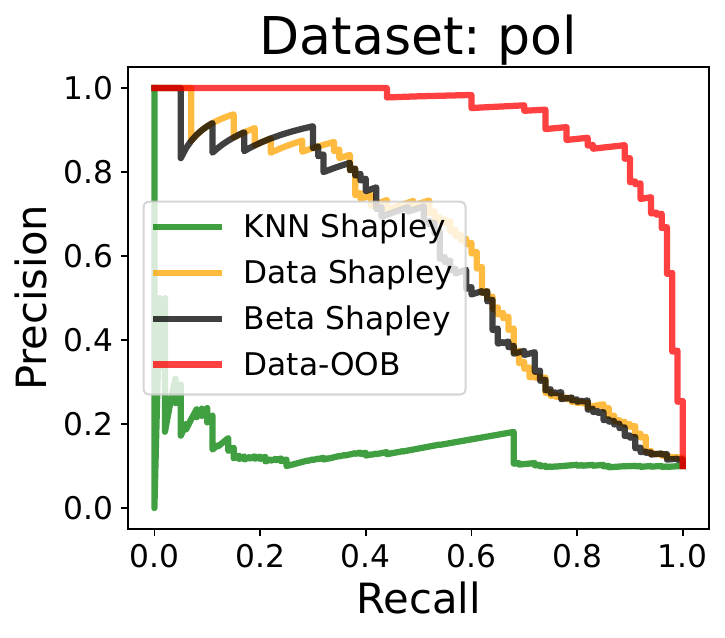}
    \includegraphics[width=0.24\textwidth]{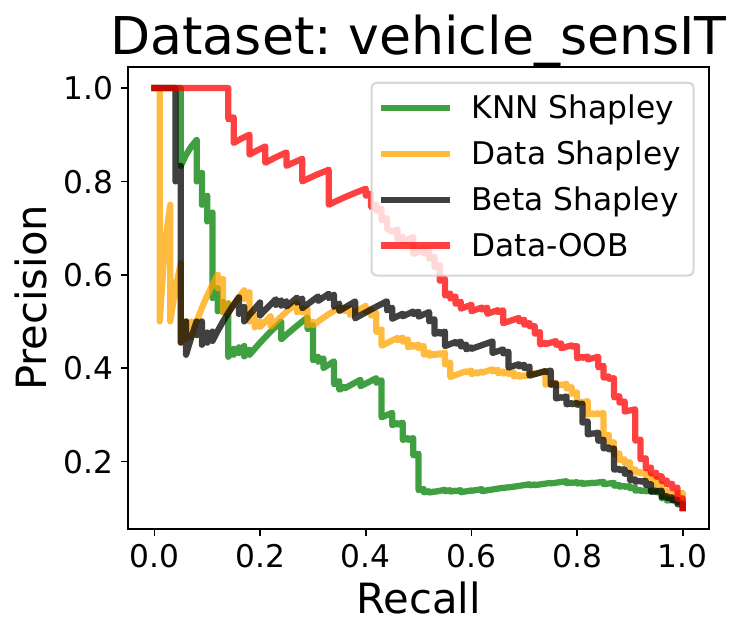}
    \includegraphics[width=0.24\textwidth]{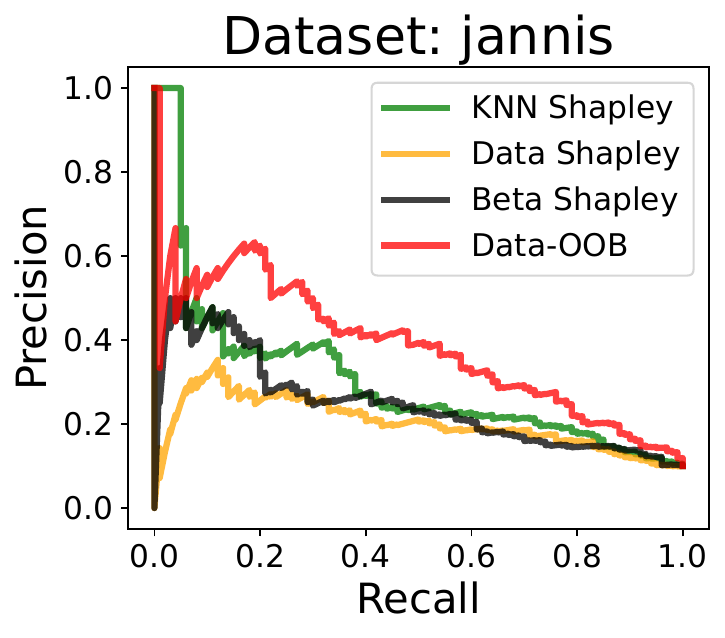}
    \includegraphics[width=0.24\textwidth]{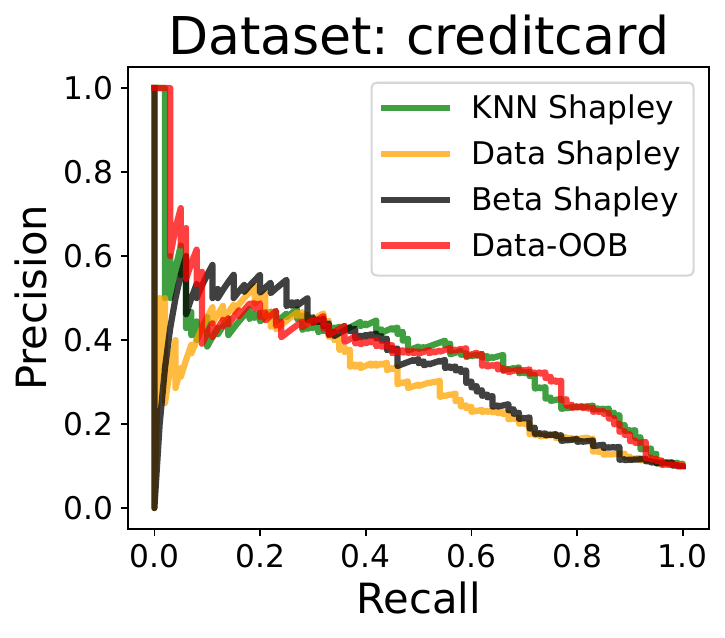}\\
    \includegraphics[width=0.24\textwidth]{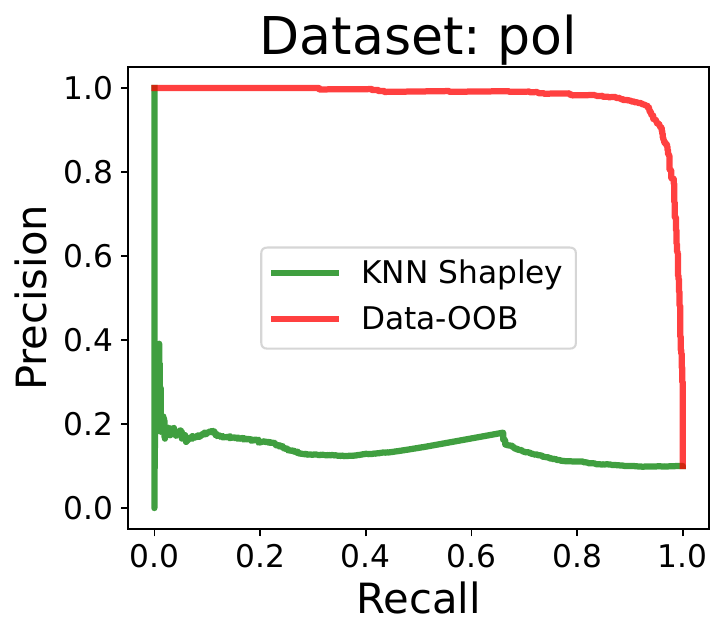}
    \includegraphics[width=0.24\textwidth]{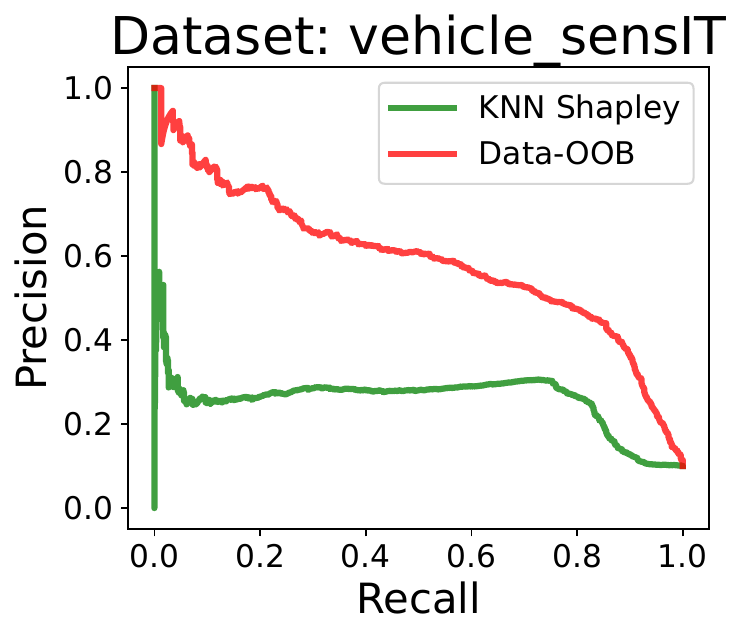}
    \includegraphics[width=0.24\textwidth]{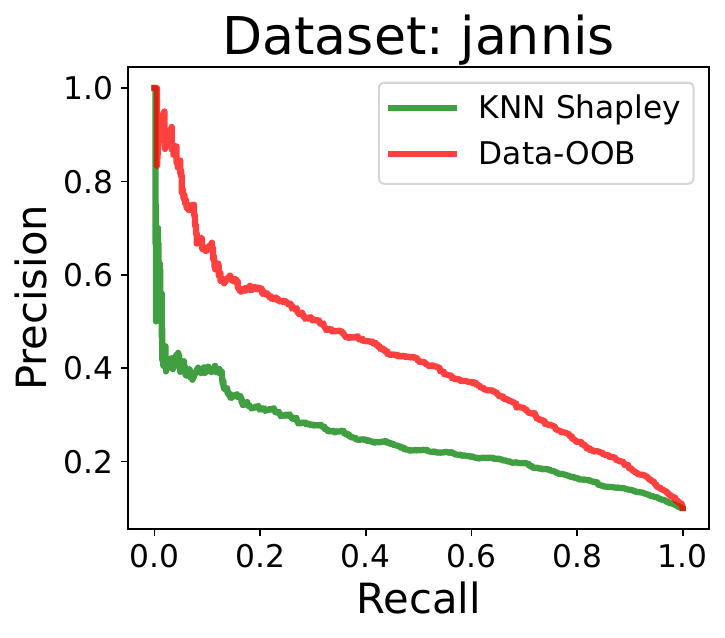}
    \includegraphics[width=0.24\textwidth]{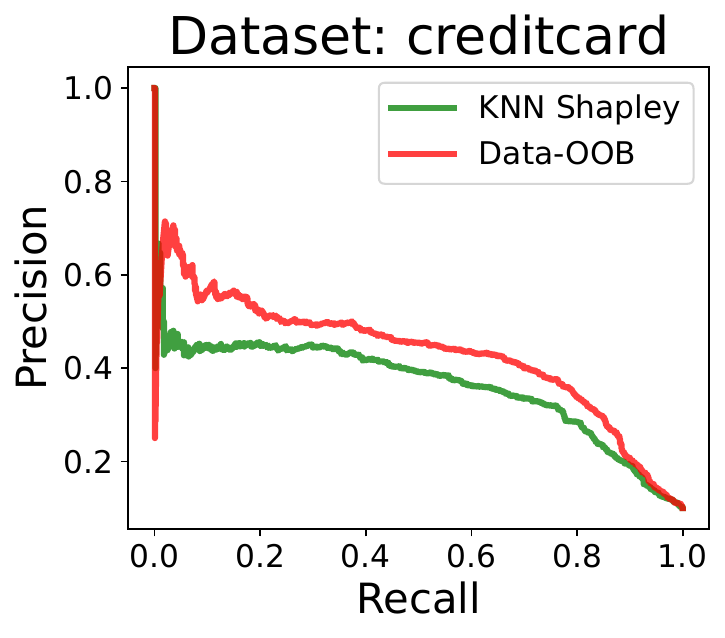}
    \vspace{-0.05in}
    \caption{Precision-recall curves of different data valuation methods on the four datasets when (top) $n=1000$ and (bottom) $n=10000$. The larger area under the curve is, the better method is. The proposed method shows superior or comparable identification performance in various settings. Additional results using different datasets are provided in Appendix~\ref{app:additional_mislabeled_data_results}.} 
    \label{fig:precision_recall_curves}
\end{figure*}

\begin{table*}[t]
    \centering
    \resizebox{\textwidth}{!}{
    \begin{tabular}{l|ccccc|cccccccc}
        \toprule 
         \multirow{2}{*}{Dataset} & \multicolumn{5}{c}{$n=1000$} & \multicolumn{3}{c}{$n=10000$} \\ 
          & \texttt{KNN Shapley} & \texttt{Data Shapley} & \texttt{Beta Shapley} & \texttt{AME} & \texttt{Data-OOB} & \texttt{KNN Shapley} & \texttt{AME} & \texttt{Data-OOB} \\ 
         \midrule 
         pol & $0.28 \pm 0.003$ & $0.50 \pm 0.011$ & $0.46 \pm 0.010$ & $0.09 \pm 0.009$ & $\mathbf{0.73 \pm 0.004}$ & $0.28 \pm 0.000$ & $0.10 \pm 0.012$ & $\mathbf{0.88 \pm 0.000}$ \\ 
         jannis & $0.25 \pm 0.004$ & $0.23 \pm 0.003$ & $0.24 \pm 0.003$ & $0.09 \pm 0.012$ & $\mathbf{0.30 \pm 0.001}$ & $0.28 \pm 0.001$ & $0.06 \pm 0.012$ & $\mathbf{0.33 \pm 0.000}$ \\ 
         lawschool & $0.45 \pm 0.014$ & $0.94 \pm 0.003$ & $0.94 \pm 0.003$ & $0.10 \pm 0.009$ & $\mathbf{0.96 \pm 0.002}$ & $0.39 \pm 0.005$ & $0.08 \pm 0.012$ & $\mathbf{0.95 \pm 0.000}$ \\ 
         fried & $0.28 \pm 0.005$ & $0.32 \pm 0.003$ & $0.32 \pm 0.004$ & $0.09 \pm 0.011$ & $\mathbf{0.44 \pm 0.004}$ & $0.35 \pm 0.001$ & $0.08 \pm 0.012$ & $\mathbf{0.54 \pm 0.001}$ \\ 
         vehicle\_sensIT & $0.20 \pm 0.004$ & $0.37 \pm 0.006$ & $0.39 \pm 0.006$ & $0.07 \pm 0.011$ & $\mathbf{0.49 \pm 0.004}$ & $0.21 \pm 0.004$ & $0.09 \pm 0.012$ & $\mathbf{0.52 \pm 0.001}$ \\ 
         electricity & $0.26 \pm 0.006$ & $0.32 \pm 0.004$ & $0.34 \pm 0.004$ & $0.08 \pm 0.010$ & $\mathbf{0.35 \pm 0.002}$ & $0.29 \pm 0.001$ & $0.08 \pm 0.012$ & $\mathbf{0.43 \pm 0.001}$ \\ 
         2dplanes & $0.30 \pm 0.007$ & $0.57 \pm 0.006$ & $0.54 \pm 0.006$ & $0.10 \pm 0.009$ & $\mathbf{0.58 \pm 0.004}$ & $0.42 \pm 0.004$ & $0.10 \pm 0.012$ & $\mathbf{0.61 \pm 0.001}$ \\ 
         creditcard & $\mathbf{0.43 \pm 0.004}$ & $0.36 \pm 0.006$ & $\mathbf{0.43 \pm 0.005}$ & $0.09 \pm 0.011$ & $0.41 \pm 0.003$ & $0.42 \pm 0.002$ & $0.12 \pm 0.012$ & $\mathbf{0.44 \pm 0.001}$ \\ 
         covertype & $\mathbf{0.51 \pm 0.021}$ & $0.37 \pm 0.004$ & $0.41 \pm 0.003$ & $0.12 \pm 0.011$ & $0.40 \pm 0.002$ & $\mathbf{0.66 \pm 0.003}$ & $0.08 \pm 0.012$ & $0.47 \pm 0.001$ \\ 
         nomao & $0.47 \pm 0.013$ & $0.65 \pm 0.005$ & $\mathbf{0.66 \pm 0.005}$ & $0.08 \pm 0.009$ & $0.65 \pm 0.004$ & $0.51 \pm 0.012$ & $0.09 \pm 0.012$ & $\mathbf{0.75 \pm 0.001}$ \\ 
         webdata\_wXa & $0.39 \pm 0.003$ & $0.38 \pm 0.006$ & $\mathbf{0.43 \pm 0.006}$ & $0.07 \pm 0.010$ & $0.40 \pm 0.003$ & $0.38 \pm 0.000$ & $0.12 \pm 0.011$ & $\mathbf{0.41 \pm 0.001}$ \\ 
         MiniBooNE & $0.33 \pm 0.008$ & $0.40 \pm 0.006$ & $0.41 \pm 0.006$ & $0.09 \pm 0.011$ & $\mathbf{0.54 \pm 0.004}$ & $0.41 \pm 0.003$ & $0.09 \pm 0.012$ & $\mathbf{0.63 \pm 0.001}$ \\ 
         \bottomrule 
    \end{tabular}
    }
    \vspace{-0.05in}
    \caption{F1-score of different data valuation methods on the twelve datasets when (left) $n=1000$ and (right) $n=10000$. The average and standard error of the F1-score based on 50 independent experiments are denoted by `average$\pm$standard error'. Bold numbers denote the best method. In almost all situations, the proposed \texttt{Data-OOB} outperforms other methods in detecting mislabeled data.}
    \vspace{-0.05in}
    \label{tab:mislabeled_data_detection}
\end{table*}

Since mislabeled data often negatively affect the model performance, it is desirable to assign low values to these data points. To see the detection ability of \texttt{Data-OOB}, we conduct mislabeled data detection experiment. We randomly choose 10\% of the entire data points and change its label to one of other labels. We first compute data values as if the contaminated dataset is the original dataset, and then we evaluate the precision and the recall of data valuation methods. Note that every method is not provided with an annotation about which data point is mislabeled. 

Figure~\ref{fig:precision_recall_curves} compares the precision-recall curves of different data valuation methods. \texttt{AME} is not displayed because it assigns the exactly zero value for most data points, resulting in meaningless precision and recall values. \texttt{Data-OOB} shows better or comparable performance than existing marginal contribution-based methods in various settings. Additional results using different datasets are provided in Appendix~\ref{app:additional_mislabeled_data_results}, where \texttt{Data-OOB} consistently shows competitive performance over \texttt{Data Shapley}, \texttt{Beta Shapley}, and \texttt{KNN Shapley}.

\begin{figure*}[t]
    \centering
    \includegraphics[width=0.235\textwidth]{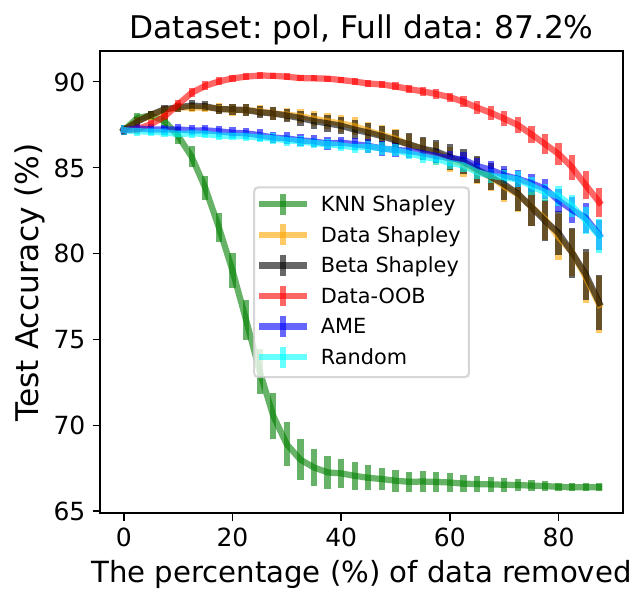}
    \includegraphics[width=0.245\textwidth]{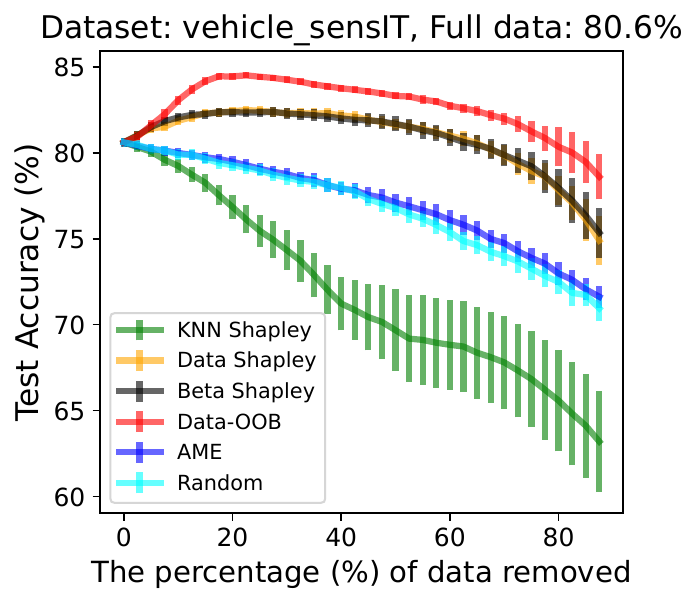}
    \includegraphics[width=0.235\textwidth]{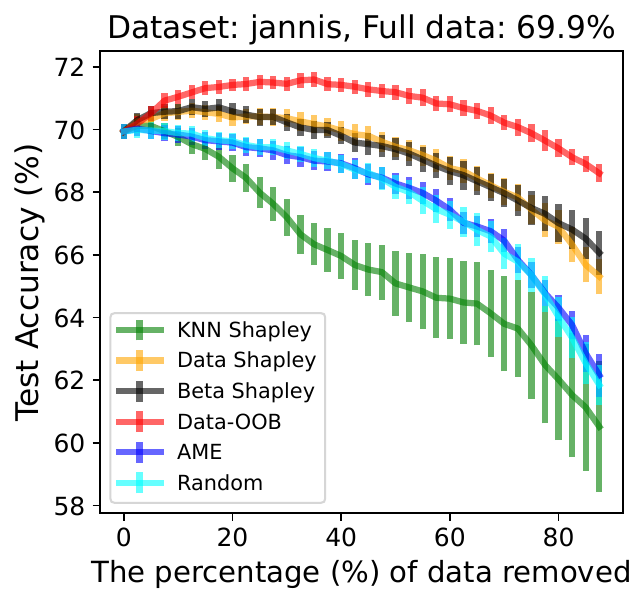}
    \includegraphics[width=0.245\textwidth]{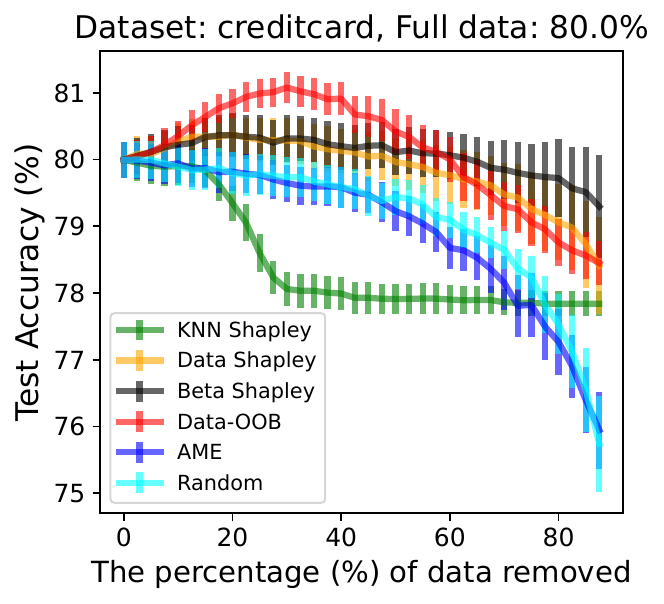}\\
    \includegraphics[width=0.235\textwidth]{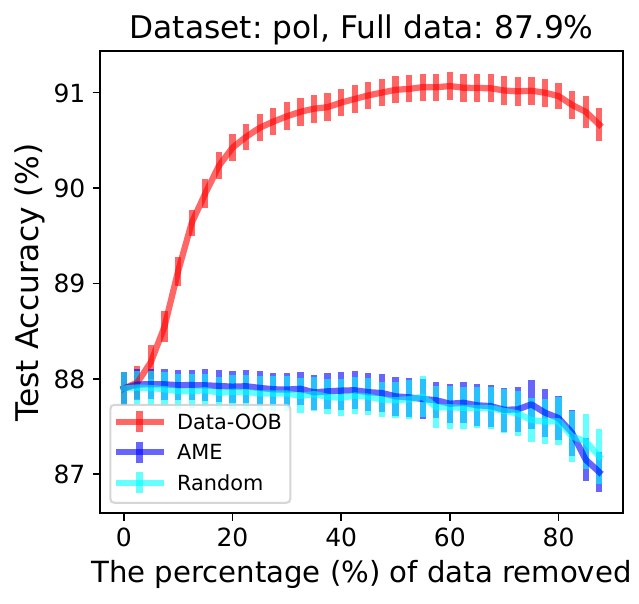}
    \includegraphics[width=0.245\textwidth]{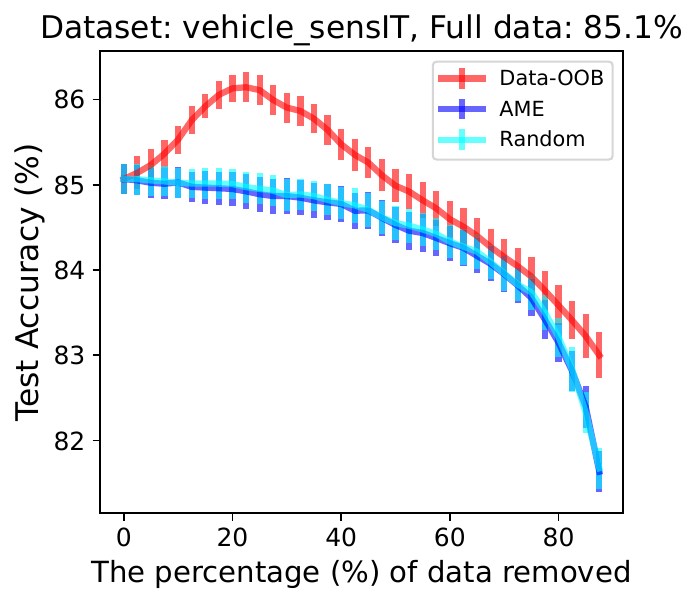}
    \includegraphics[width=0.235\textwidth]{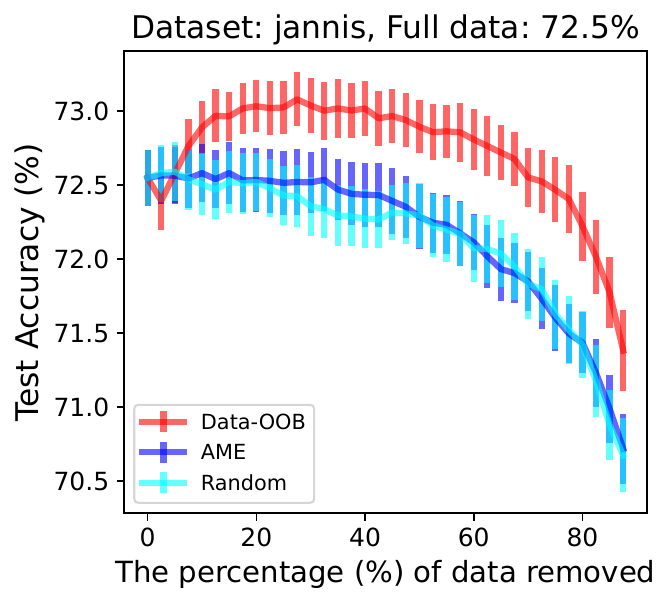}
    \includegraphics[width=0.245\textwidth]{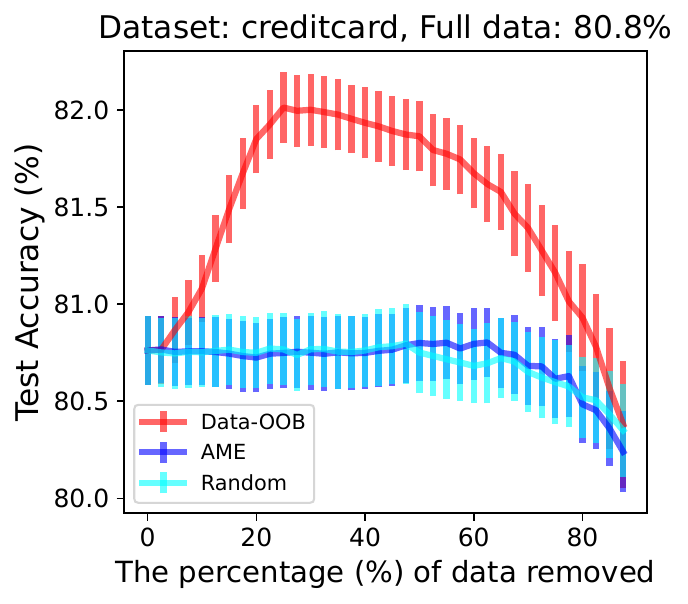}
    \vspace{-0.05in}
    \caption{Test accuracy curves as a function of the percentage of data removed. We consider the four datasets when (top) $n=1000$ and (bottom) $n=10000$. We remove data points one by one starting from the lowest to the largest. Higher curves indicate better performance for data valuation. The error bar indicates a 95\% confidence interval based on 50 independent experiments. Additional results using 8 different datasets are provided in Appendix~\ref{app:additional_point_removal_results}.}
    \label{fig:point_removal_experiment}
    \vspace{-0.05in}
\end{figure*}

We further assess the detection ability of different data valuation methods. Following the mislabeled data detection task in \citet{kwon2022beta}, we apply the K-means algorithm to data values and divide data points into two clusters. \citep{arthur2007k}. We regard data points in a cluster with a lower mean as the prediction for mislabeled data points. Then, the F1-score is evaluated by comparing the prediction with its actual annotations. Table~\ref{tab:mislabeled_data_detection} shows the F1-score of different data valuation methods for the twelve classification datasets. Overall, \texttt{Data-OOB} significantly outperforms other state-of-the-art methods. In particular, when dataset is `pol' and $n=10000$, \texttt{Data-OOB} achieves $3.1$ and $8.7$ times greater F1-score than \texttt{KNN Shapley} and \texttt{AME}, respectively. As noted by \citet{lin2022measuring}, the F1-score for \texttt{AME} can be improved if the Model-X Knockoff procedure is incorporated \citep{candes2018panning}. However, it requires additional training LASSO models with dummy variables, resulting in extra computational costs. We demonstrate that \texttt{Data-OOB} shows strong performance in detecting mislabeled data points without such procedures. % that may not be applied to other downstream tasks.

\subsection{Point removal experiment}
\label{sec:point_removal_experiment}
Data valuation methods can be useful in identifying a small set of the training dataset that is helpful or harmful to a model's performance. To see this, we conduct the point removal experiment. Following \citet{ghorbani2019}, we remove data points from the entire dataset one by one starting from lowest to largest. At each time a datum is removed, we fit a logistic regression model with the remaining data and evaluate the test accuracy of the logistic regression model. We include the random removal as one of baseline methods, and use the same contaminated datasets in Section~\ref{sec:mislabeled_data_experiment}. All the test accuracy results are evaluated on the fixed holdout dataset with $3000$ data points. 

Figure~\ref{fig:point_removal_experiment} shows the test accuracy curves on the four datasets when $n \in \{1000, 10000\}$. Overall, \texttt{Data-OOB} achieves significantly better test accuracy than other baseline methods in various settings, showing a steeper increase in the first 20\% of data removal. 
We suggest that this increase in performance is due to the mislabeled data detection performance as mislabeled data points are likely to be removed first in the first 20\% of data removal. %In this respect, it is closely related to Section~\ref{sec:mislabeled_data_experiment}, and our method achieves competitive model performance.

After the 80\% of data removal interval, \textit{i.e.}, with top 20\% helpful data points, the proposed method maintains the test accuracy similar to the level of the entire dataset. For instance, when the dataset is `jannis' and $n=10000$, the test accuracy after removing 80\% of the dataset is 72.2\%, which is only 0.3\% degradation compared to the entire data set. Surprisingly, in case of the `pol' dataset with $n=10000$, the accuracy is improved to 90.9\% from 87.9\% by removing unhelpful data points. Our experiments demonstrate that \texttt{Data-OOB} can be a practical solution for not only selecting detrimental data but finding a pivotal set that can maintain or improve test accuracy with a smaller dataset.

As for the \texttt{AME}, we find that it performs similarly to the random removal. This is because the LASSO model in \texttt{AME} assigns exactly zero values for most of the data points, leading to a similar behavior to the random selection. The sparsity is usually regarded as a desired property in high dimensional data analysis, but it shows that it may not be critical in data valuation problems. As for the \texttt{KNN Shapley}, it shows the worst performance when $n=1000$ and $n=10000$, and the case $n=10000$ is excluded from Figure~\ref{fig:point_removal_experiment} for a better presentation. This poor performance may result from the $k$-nearest neighbors algorithm that is essential for \texttt{KNN Shapley}. The $k$-nearest neighbors algorithm only uses data points in a local neighborhood, and as a result, it can fail to capture the information characterized by the data distribution (\textit{i.e.}, margin from a classification boundary).

To further investigate the effect of data removal on the model performance, we illustrate a distribution of the remaining data points when 20\%, 50\%, or 80\% of data points are removed. We compare \texttt{Data-OOB} with \texttt{KNN Shapley } and \texttt{AME}. We use the `MiniBooNE' dataset with $n=1000$. As its input dimension is $50$, we display data points using their first two principal component scores. % The variance explained by the first two principal components is XXX\%.

\begin{figure*}[t]
    \centering
    \includegraphics[width=0.98\textwidth]{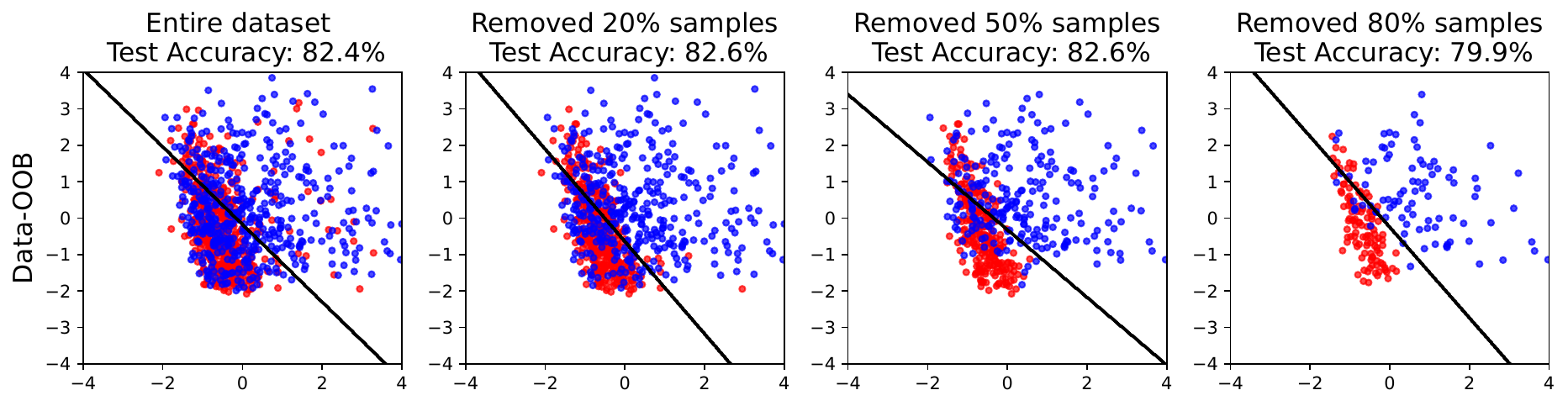}
    \includegraphics[width=0.98\textwidth]{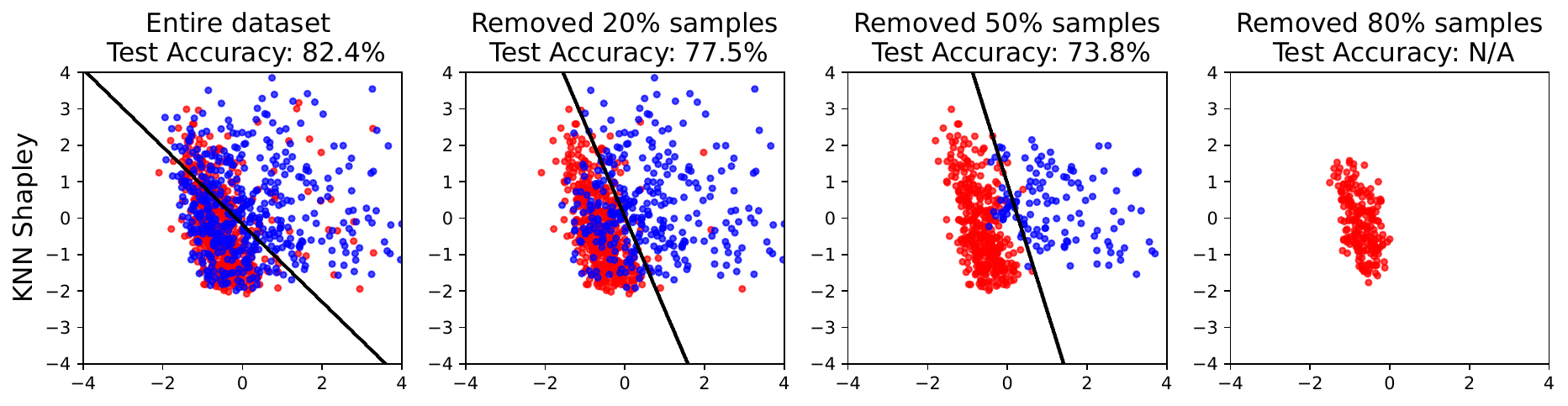} 
    \includegraphics[width=0.98\textwidth]{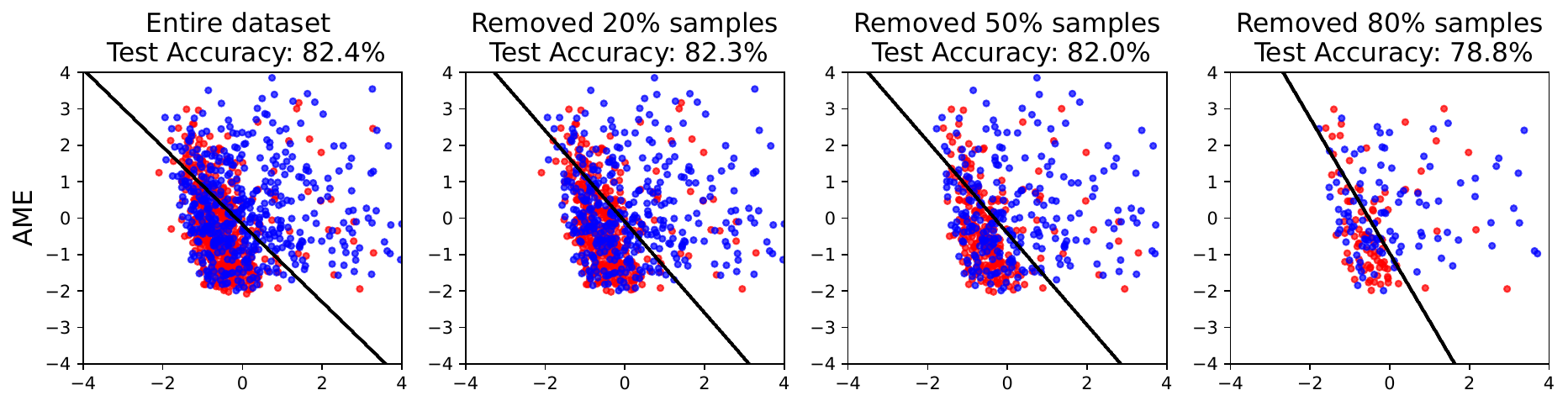}
    \vspace{-0.1in}
    \caption{Distribution after data removal for (top) \texttt{Data-OOB}, (middle) \texttt{KNN Shapley}, and (bottom) \texttt{AME}. We remove data points one by one starting from the lowest to the largest. We illustrate data points with their first two principal components. The color indicates a class and the black solid line indicates a classifier boundary obtained by training the remaining dataset. We intentionally repeat the three figures in the first column to ease the comparison. The proposed method shows a better capability in finding beneficial data points.} 
    \label{fig:distribution_of_subsets}
    \vspace{-0.075in}
\end{figure*}

Figure~\ref{fig:distribution_of_subsets} shows the four different snapshots, namely 0\%, 20\%, 50\%, and 80\% of data removal. \texttt{Data-OOB} shows an increased test accuracy even after 50\% of data removal by effectively removing unhelpful data points. When 80\% of data are removed, it clearly shows a region of each class, giving an intuitive classifier boundary. The test accuracy for \texttt{KNN Shapley} after 80\% of data removal is not measured as there are no blue class data points in top 20\%. This is anticipated in that \texttt{KNN Shapley} overly focuses on the local neighbors and tends to assign large values if there is a homogeneous local neighbors. \texttt{AME} shows a better test accuracy than \texttt{KNN Shapley}, but it almost randomly removes data points due to sparse data values. As a result, it does not give meaningful insights into class regions.
In Appendix~\ref{app:distribution_after_data_removal}, we show additional experiments with the datasets `electricity' and `fried'. \texttt{Data-OOB} consistently identifies informative data points, showing a better capability in finding beneficial data points than \texttt{KNN Shapley} and \texttt{AME}.

We emphasize that \texttt{Data-OOB} is insensitive to different choices of the number of weak learners $B$. We conduct a mislabeled data detection experiment with $B \in \{400, 800, 3200\}$, showing that the F1-score for \texttt{Data-OOB} remains stable. It shows that our experimental results continue to hold with a smaller $B$. We provide this result in Appendix~\ref{app:robustness}.

\section{Related Works}
\label{sec:related_works}

\paragraph{Bagging}
Bootstrap aggregation, which is also known as bagging, is an ensemble technique that trains multiple weak learners where each learner is trained using a bootstrap dataset \citep{breiman1996bagging}. One popular and powerful bagging model is the random forest in which multiple numbers of decision trees are trained with a randomly selected set of features \citep{breiman2001random, wager2014confidence, athey2019generalized}. While the primary usage of bagging is to improve a model's performance by decreasing the variance of its predictions, the proposed Data-OOB presents a distinct application of bagging.

\paragraph{Marginal contribution-based methods in machine learning} Marginal contribution-based methods have been studied and applied to various machine learning problems, for instance, feature attribution problems \citep{lundberg2017unified, covert2021explaining, kwon2022weightedshap}, model explanation \citep{stier2018analysing, ghorbani2020neuron}, collaborative learning \citep{sim2020collaborative, xu2021gradient}, and federated learning \citep{wang2019interpret, wang2020principled}. The Shapley value is one of the most widely used marginal contribution-based methods, and many alternative approaches have been studied by relaxing some of the underlying fair division axioms \citep{yan2021if, kwon2022beta, wang2022data, rozemberczki2022shapley}. Alternatively, there have been approaches that are independent of marginal contributions. In the data valuation literature, for instance, \citet{yoon2020data} proposed a data value estimator model using reinforcement learning and \citet{ilyas2022datamodels} proposed datamodels that capture the influence via predicting a model's prediction as \citet{lin2022measuring}.

\section{Concluding Remarks}
\label{sec:conclusion}

In this paper, we propose Data-OOB that is suitable for any tabular machine learning datasets as it is easy to train a random forest model on such datasets. With comprehensive numerical experiments, we demonstrate that Data-OOB is significantly powerful in identifying helpful and harmful data points for model training. Our method does not require additional validation points and is computationally efficient by reusing trained weak learners. Data-OOB is statistically interpretable, showing it has the order consistency with the infinitesimal jackknife influence function.

While Data-OOB has shown promising results in various classification datasets, there are several limitations and it opens several future avenues of research. One potential extension of Data-OOB is to leverage weak learners in boosting models instead of bagging models. We find that boosting models should be treated differently from a regular bagging model. This is because a weak learner in boosting predicts the residuals obtained from the previous optimization steps, not predicting the ground truth labels. In other words, a weak learner in boosting is sequentially dependent on other weak learners, making a direct application of Data-OOB challenging in downstream machine learning tasks. We believe computing data values with a trained boosting model could be very influential as boosting often performs better than a random forest in practice.

One potential caveat is that Data-OOB can assign erroneously high values for detrimental data points if there are many duplicates. This is because when there are multiple duplicate data, the OOB estimate becomes similar to the training accuracy, not the test accuracy. We believe a simple removal of duplicates can address this issue, but we encourage the community to develop a more principled method for duplicate data.

\section*{Acknowledgements}
The authors would like to thank all anonymous reviewers for their helpful comments. We also would like to thank Mert Yuksekgonul and Young-geun Kim for their constructive feedback.

\bibliography{ref}
\bibliographystyle{icml2023}

%%%%%%%%%%%%%%%%%%%%%%%%%%%%%%%%%%%%%%%%%%%%%%%%%%%%%%%%%%%%%%%%%%%%%%%%%%%%%%%
%%%%%%%%%%%%%%%%%%%%%%%%%%%%%%%%%%%%%%%%%%%%%%%%%%%%%%%%%%%%%%%%%%%%%%%%%%%%%%%
% APPENDIX
%%%%%%%%%%%%%%%%%%%%%%%%%%%%%%%%%%%%%%%%%%%%%%%%%%%%%%%%%%%%%%%%%%%%%%%%%%%%%%%
%%%%%%%%%%%%%%%%%%%%%%%%%%%%%%%%%%%%%%%%%%%%%%%%%%%%%%%%%%%%%%%%%%%%%%%%%%%%%%%
\newpage
\appendix
\onecolumn

\section{Implementation Details}
\label{app:implementation_details}
In this section, we provide implementation details. Our Python-based implementation codes are publicly available at \url{https://github.com/ykwon0407/dataoob}.

\paragraph{Datasets}
We use 12 classification datasets in Section~\ref{sec:experiment}. The `covertype' dataset is downloaded via the Python package `scikit-learn' \citep{Pedregosa2011Scikit}, and every other dataset is downloaded from `OpenML' \citep{feurer-arxiv19a}. Table~\ref{tab:summary_of_datasets} shows a summary of classification datasets.

We apply a standard normalization procedure. Each feature is normalized to have zero mean and one standard deviation. After this preprocessing, we split it into the three datasets, namely, a training dataset, a validation dataset, and a test dataset. We evaluate the value of data in the training dataset and use the validation dataset to evaluate the utility function. Note that the proposed method does not use this validation dataset, and it essentially uses a smaller dataset. The test dataset is used for point removal experiments only when evaluating the test accuracy. The training dataset size $n$ is either $1000$ or $10000$, and the validation size is fixed to 10\% of the training sample size. The test dataset size is fixed to $3000$.

\begin{table}[h]
    \centering
    \resizebox{\textwidth}{!}{
    \begin{tabular}{l|cccccccccc}
        \toprule
        Name & Sample size & Input dimension & Number of Classes & OpenML ID & Minor class proportion \\ 
         \midrule 
         law-school-admission-bianry & 20800 & 6 & 2 & 43890 & 0.321 \\ 
         electricity & 38474 & 6 & 2 & 44080 & 0.5 \\ 
         fried & 40768 & 10 & 2 & 901 & 0.498 \\ 
         2dplanes & 40768 & 10 & 2 & 727 & 0.499 \\ 
         default-of-credit-card-clients & 30000 & 23 & 2 & 42477 & 0.221 \\ 
         pol & 15000 & 48 & 2 & 722 & 0.336 \\ 
         MiniBooNE & 72998 & 50 & 2 & 43974 & 0.5 \\ 
         jannis & 57580 & 54 & 2 & 43977 & 0.5 \\ 
         nomao & 34465 & 89 & 2 & 1486 & 0.285 \\ 
         vehicle\_sensIT & 98528 & 100 & 2 & 357 & 0.5 \\ 
         webdata\_wXa & 36974 & 123 & 2 & 350 & 0.240 \\ 
         covertype & 581012 & 54 & 7 & Scikit-learn & 0.004\\
        \bottomrule
    \end{tabular}}
    \caption{A summary of 12 classification datasets used in our experiments. We provide the dataset-specific OpenML ID in the column `OpenML ID'.}
    \label{tab:summary_of_datasets}
\end{table}

\paragraph{Hyperparameters for data valuation methods}
\begin{itemize}
    \item For \texttt{KNN Shapley}, the only hyperparameter is the number of nearest neighbors. Since there is no optimal fixed number for hyperparameter, we set it to be 10\% of the sample size $n$ motivated by \citet{jia2019b}.
    \item For \texttt{Data Shapley} and \texttt{Beta Shapley}, following \citet{kwon2022beta}, we use a Monte Carlo-based algorithm. Specifically, it consists of two stages. In the first stage, we estimate every marginal contribution and in the second stage, we compute the Shapley value or semivalues. The second stage is straightforward, so here we explain the first stage. We first randomly draw a cardinality $j$ from a discrete uniform distribution on $[n]$. Then, we uniformly draw a subset $S$ from a set of subsets with the cardinality $j$. After that, we evaluate utility $U(S)$. We construct 10 independent Monte Carlo chains for this procedure and compute the Gelman-Rubin statistics to check the convergence of a simple average of marginal contributions. For each data point, we can compute the Gelman-Rubin statistics, and we consider the maximum of these statistics across samples. We stop the algorithm if the maximum value is less than the threshold value $1.05$, which is less than a usual threshold $1.1$ \citep{gelman1995bayesian}. We use a decision tree model for the utility evaluation for a fair comparison with the proposed method. 
    \item For \texttt{AME}, we set the number of utility evaluations to be $800$. Following \citet{lin2022measuring}, we consider the same uniform distribution for constructing subsets. That is, for each $p \in \{0.2, 0.4, 0.6, 0.8\}$, we randomly generate $200$ subsets such that the probability that a datum is included in the subset is $p$. The number of utility evaluation is chosen to be same with the number of weak learners $B$ of the proposed algorithm for a fair comparison. Like \texttt{Data Shapley} and \texttt{Beta Shapley}, we use a decision tree model for the utility evaluation. As for the Lasso model, we optimize the regularization parameter using `LassoCV' in `scikit-learn' with its default parameter values.
    \item The proposed method fits a random forest model with $B=800$ decision trees using `scikit-learn'. In classification settings, we use $T(y_1, y_2) = \mathds{1}(y_1= y_2)$.
\end{itemize}

\section{Additional Experimental Results}
\label{app:additional_experiments}
In this section, we present additional experimental results using the eight classification datasets. 

\subsection{Mislabeled Data Detection}
\label{app:additional_mislabeled_data_results}
\begin{figure*}[h]
    \centering
    \begin{subfigure}
        \centering
        \includegraphics[width=0.225\textwidth]{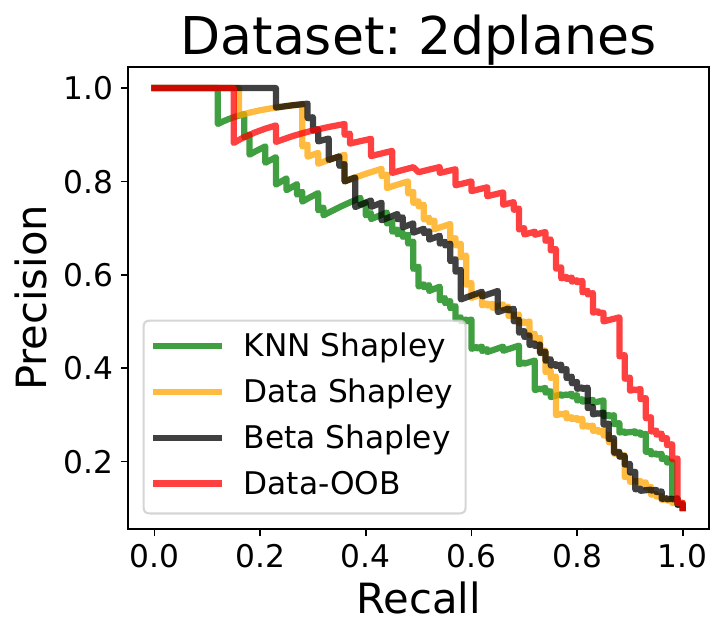}
        \includegraphics[width=0.225\textwidth]{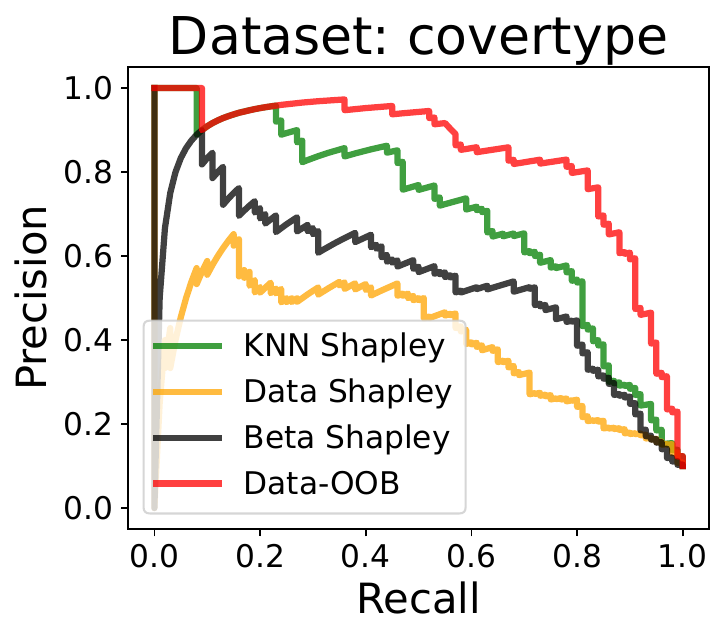}
        \includegraphics[width=0.225\textwidth]{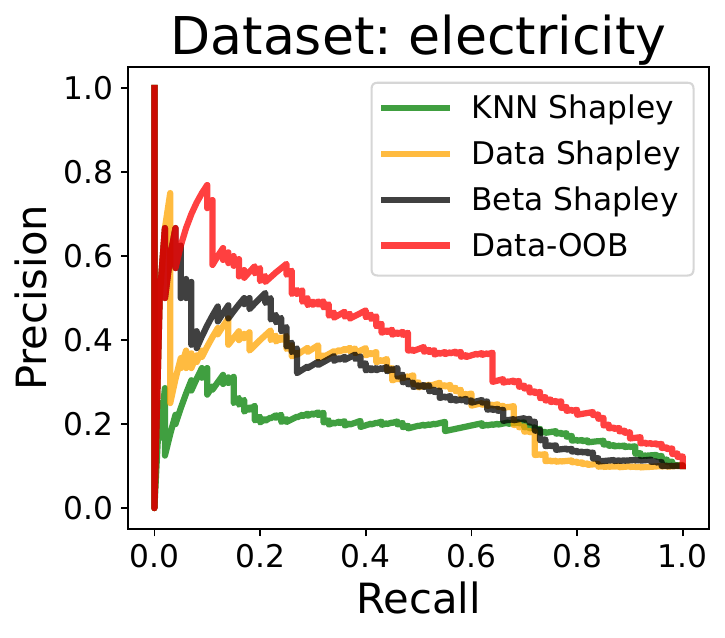}
        \includegraphics[width=0.225\textwidth]{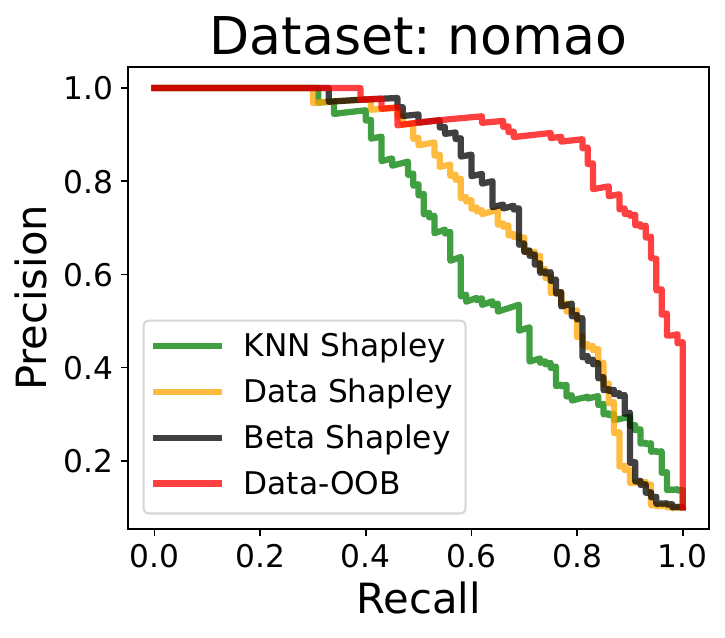}\\
        \includegraphics[width=0.225\textwidth]{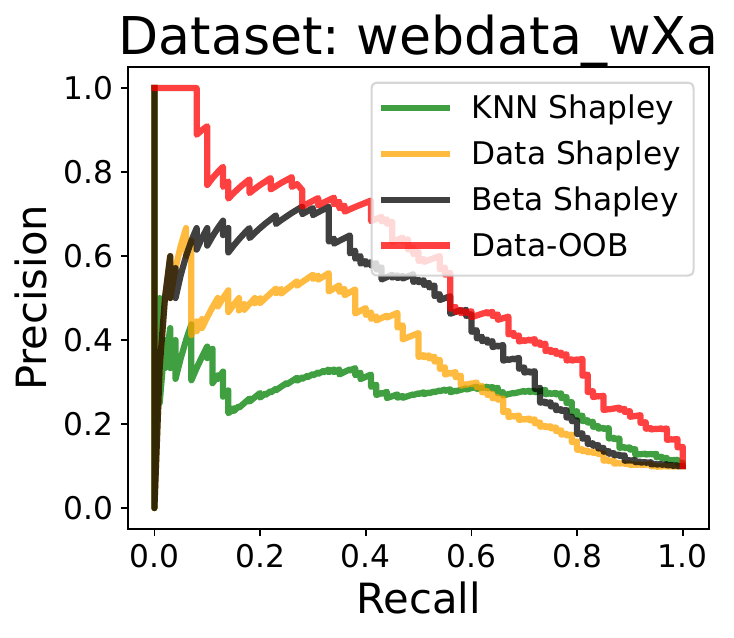}
        \includegraphics[width=0.225\textwidth]{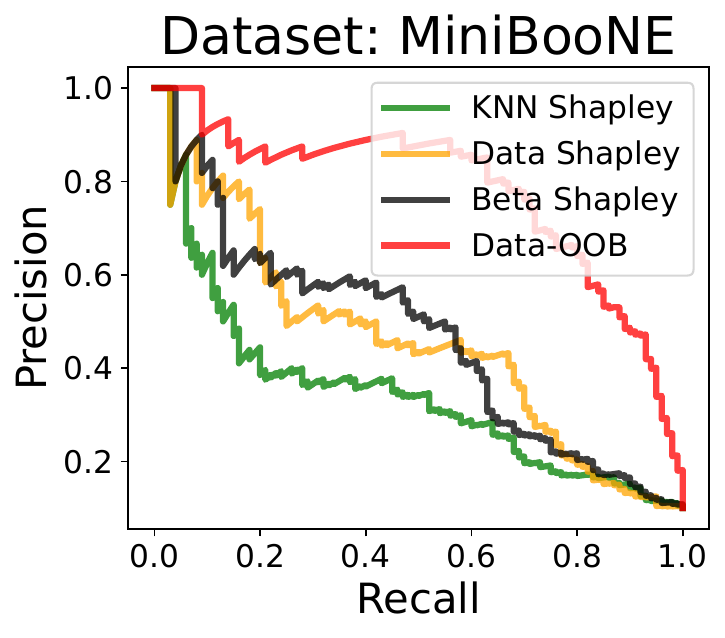}
        \includegraphics[width=0.225\textwidth]{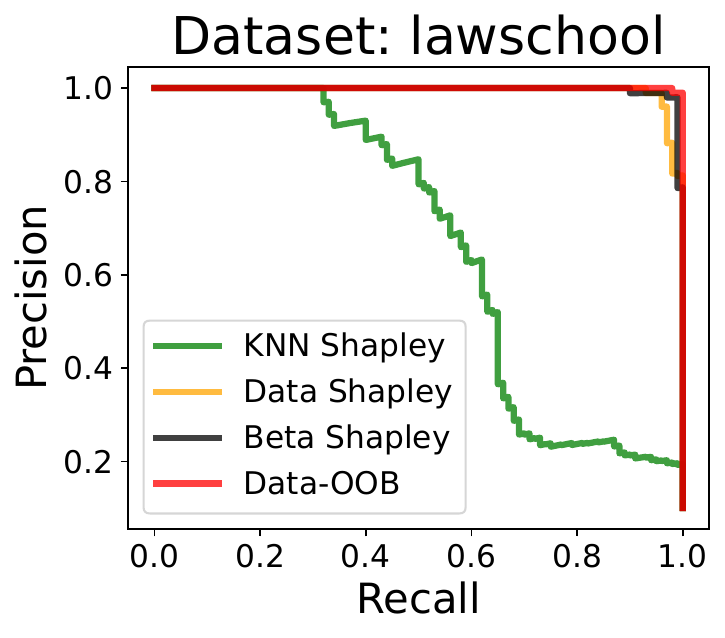}
        \includegraphics[width=0.225\textwidth]{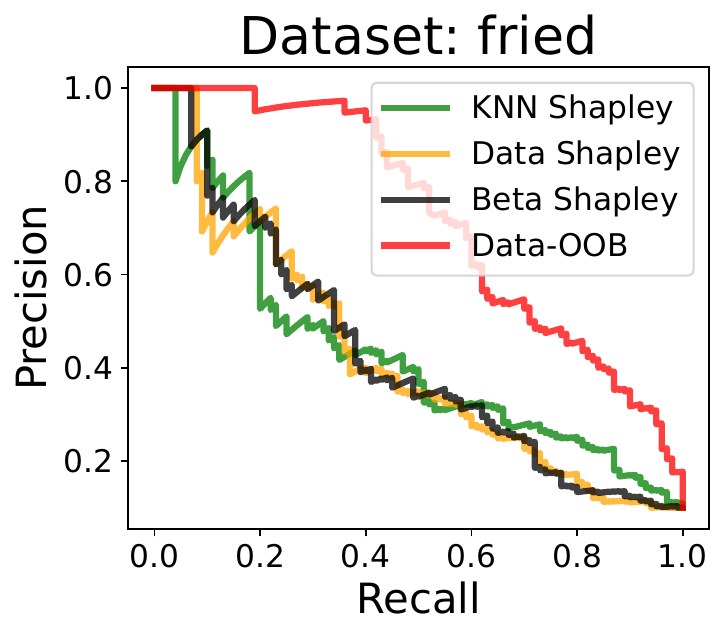}
        \caption{Precision-recall curves of different data valuation methods on the eight datasets when $n=1000$. The larger are under the curve is, the better method is.}
        \label{fig:precision_recall_curve_additional_datasets1000}
    \end{subfigure}
    \centering
    \begin{subfigure}
        \centering
        \includegraphics[width=0.225\textwidth]{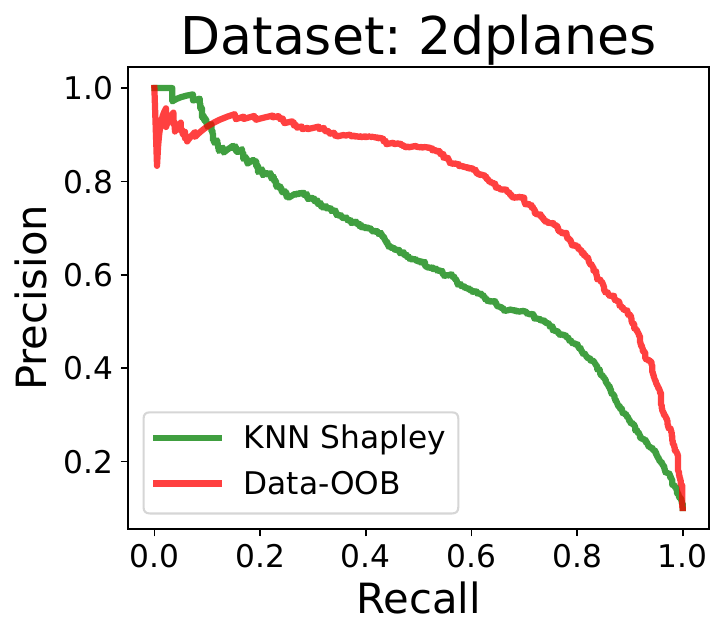}
        \includegraphics[width=0.225\textwidth]{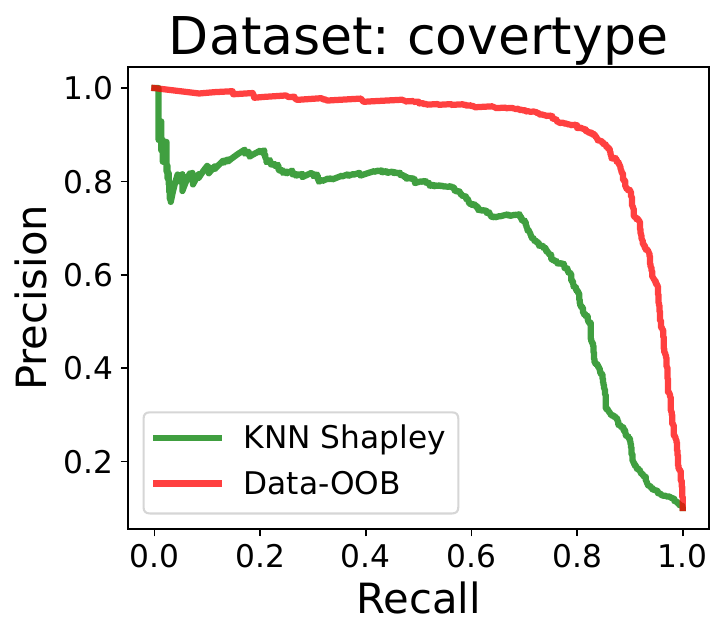}
        \includegraphics[width=0.225\textwidth]{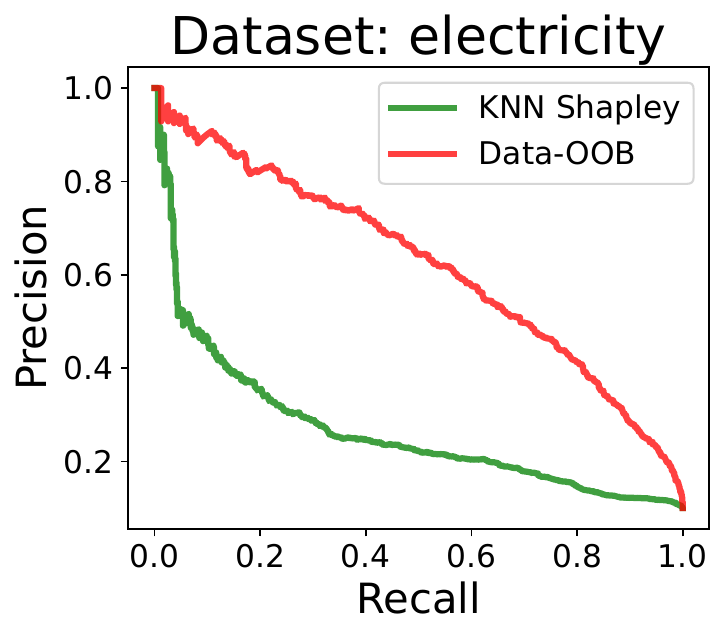}
        \includegraphics[width=0.225\textwidth]{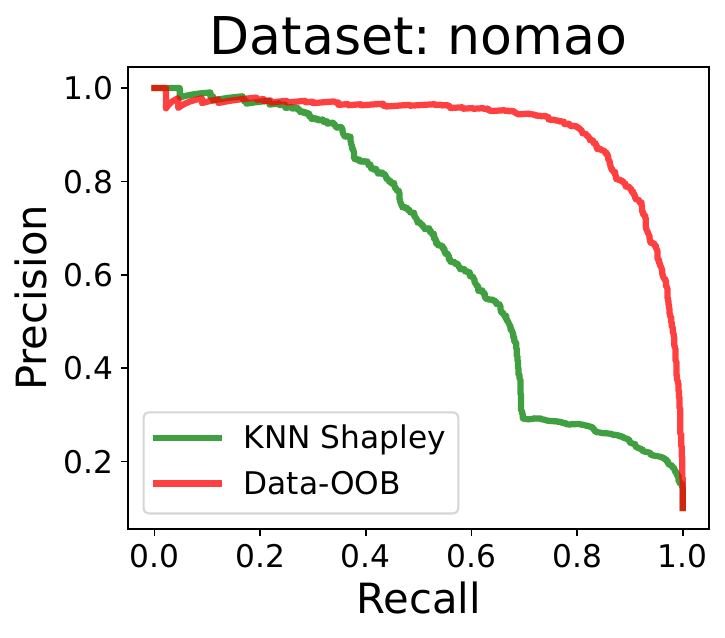}\\
        \includegraphics[width=0.225\textwidth]{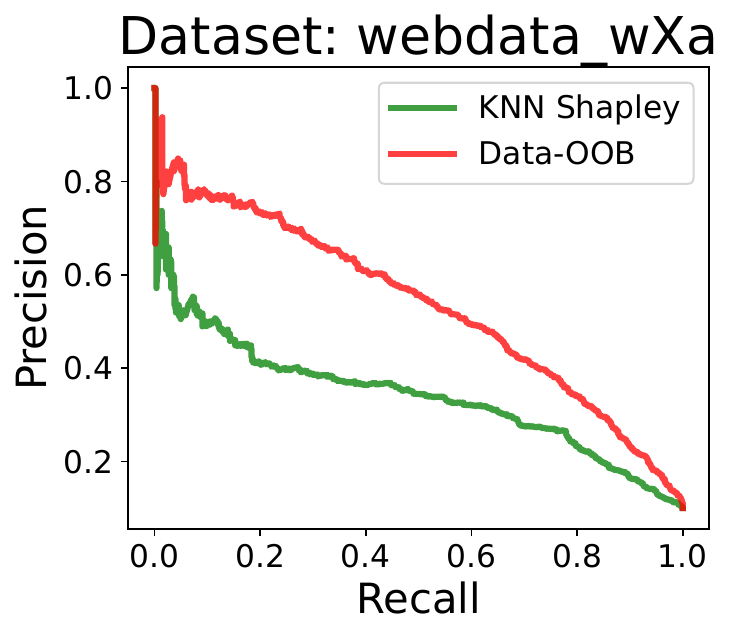}
        \includegraphics[width=0.225\textwidth]{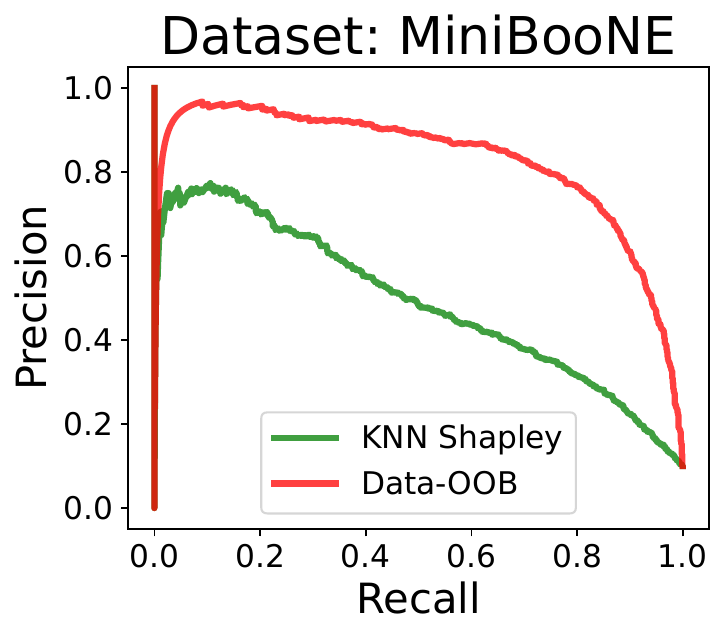}
        \includegraphics[width=0.225\textwidth]{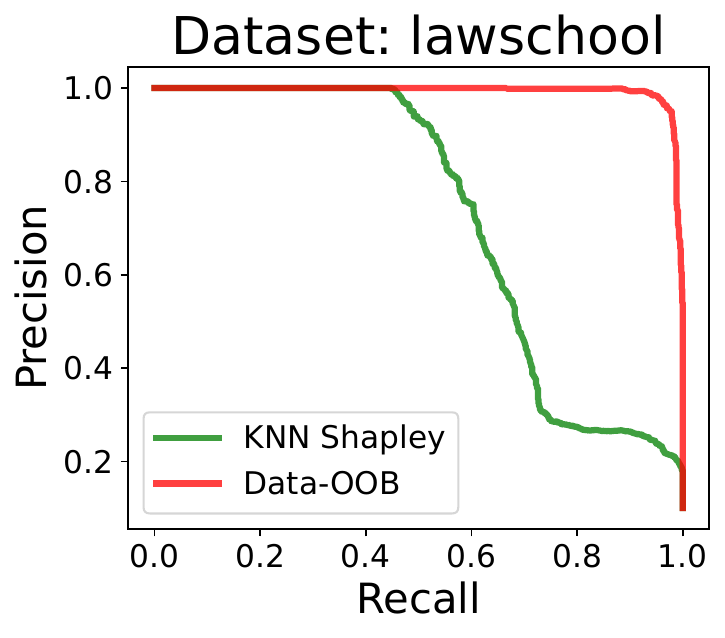}
        \includegraphics[width=0.225\textwidth]{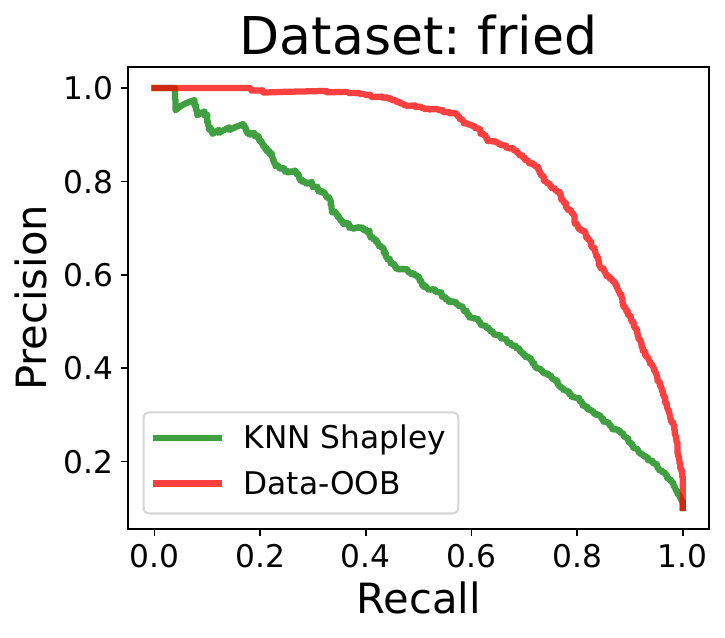}
        \caption{Precision-recall curves of different data valuation methods on the eight datasets when $n=10000$. The larger are under the curve is, the better method is.}
        \label{fig:precision_recall_curve_additional_datasets10000}
    \end{subfigure}
\end{figure*}

We provide the precision-recall curves for the eight datasets when $n \in \{1000, 10000\}$. Except for datasets, every experimental setting is exactly the same as the one used in Figure~\ref{fig:precision_recall_curves}. Figures~\ref{fig:precision_recall_curve_additional_datasets1000} and \ref{fig:precision_recall_curve_additional_datasets10000} show that the proposed method has superior or at least comparable identification performance in various settings.  

\newpage
\subsection{Point Removal Experiment}
\label{app:additional_point_removal_results}
\begin{figure*}[h]
    \centering
    \begin{subfigure}
        \centering
        \includegraphics[width=0.225\textwidth]{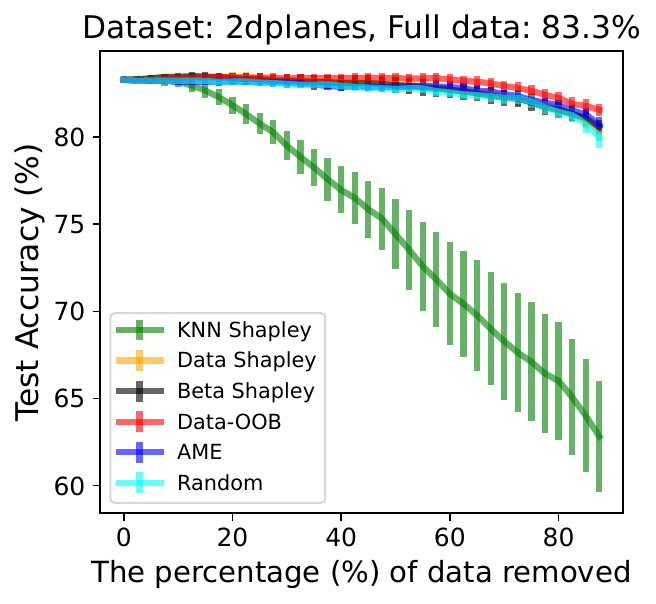}
        \includegraphics[width=0.225\textwidth]{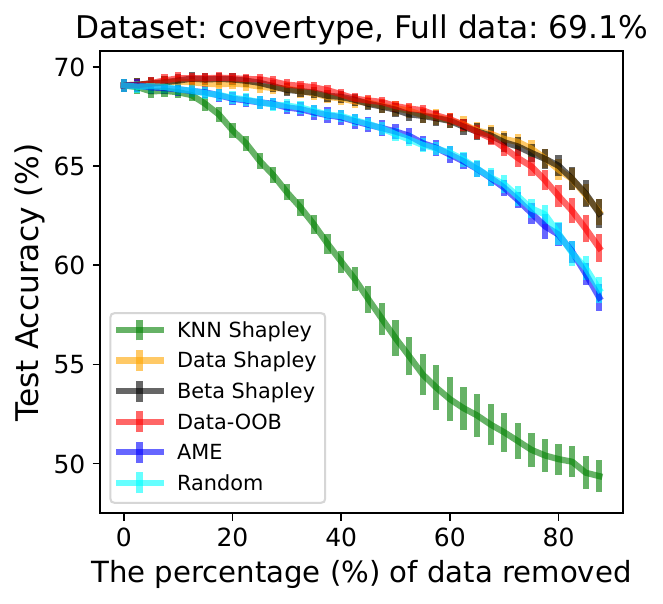}
        \includegraphics[width=0.225\textwidth]{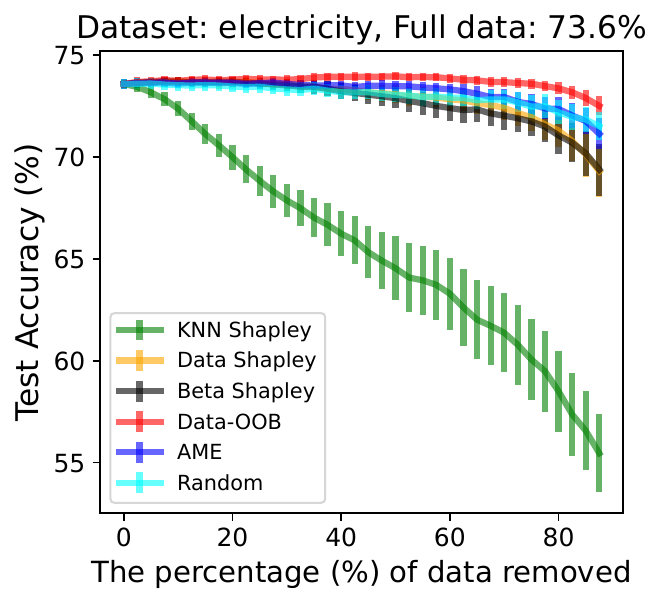}
        \includegraphics[width=0.225\textwidth]{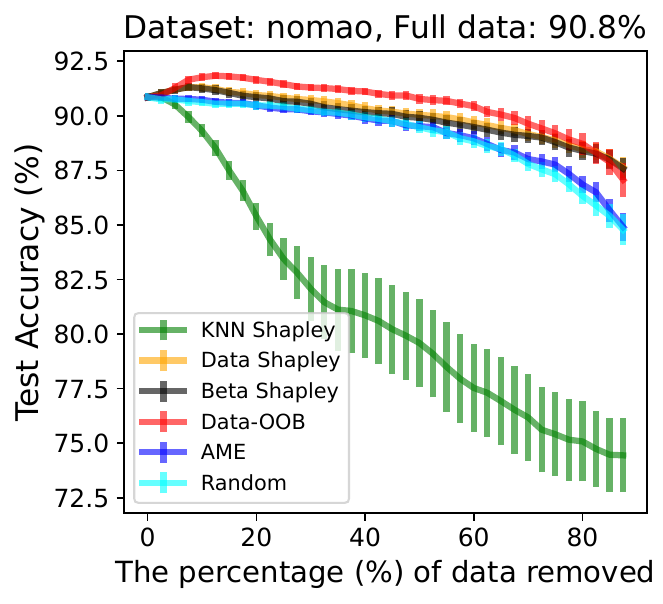}\\
        \includegraphics[width=0.225\textwidth]{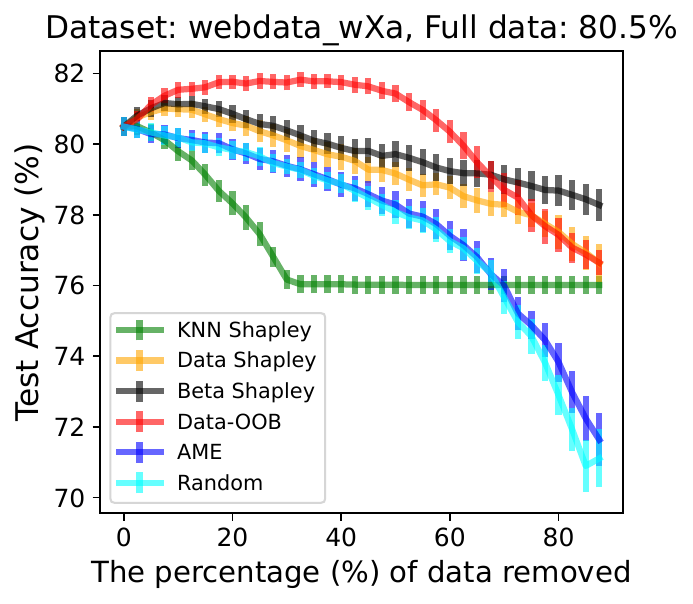}
        \includegraphics[width=0.225\textwidth]{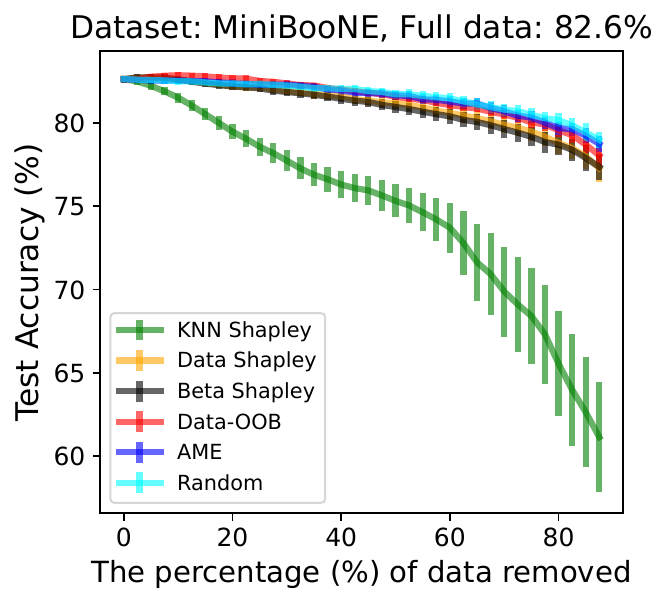}
        \includegraphics[width=0.225\textwidth]{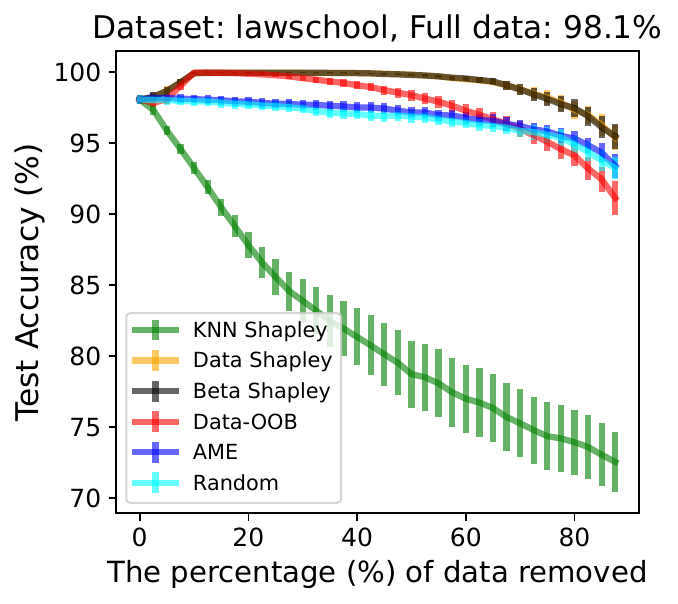}
        \includegraphics[width=0.225\textwidth]{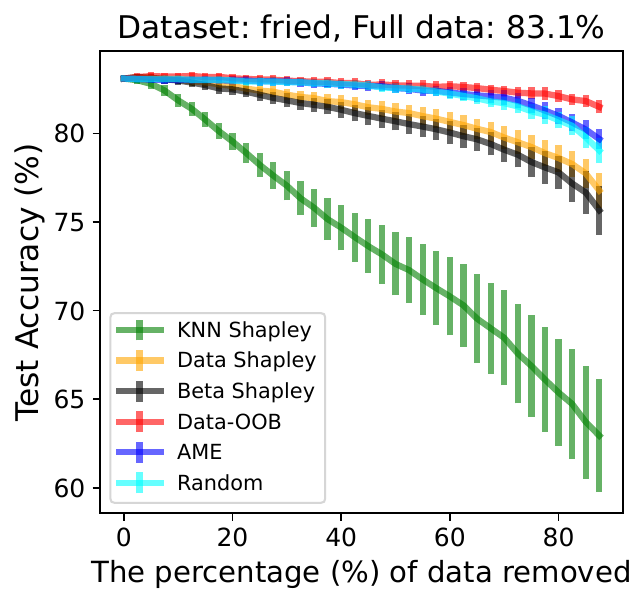}
        \caption{Test accuracy curves as a function of the percentage of data removed. We consider the four datasets when $n=1000$.}
        \label{fig:point_removal_experiment_additional_experiment1000}
    \end{subfigure}
    \centering
    \begin{subfigure}
        \centering
        \includegraphics[width=0.225\textwidth]{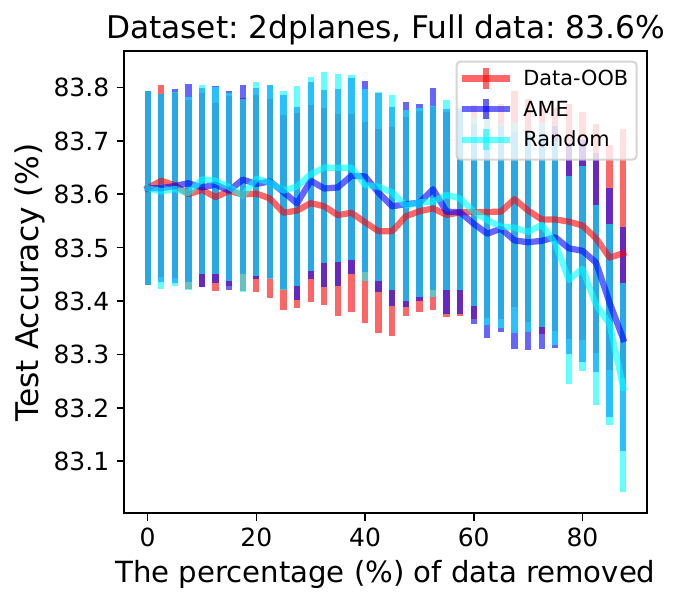}
        \includegraphics[width=0.225\textwidth]{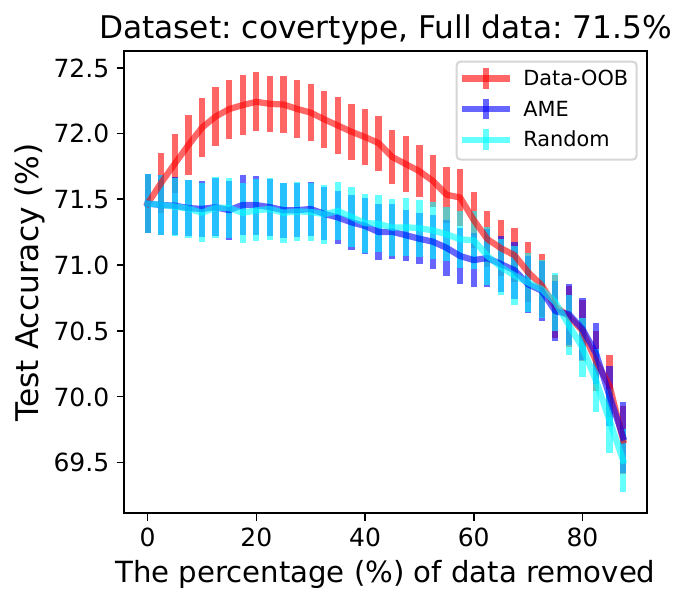}
        \includegraphics[width=0.225\textwidth]{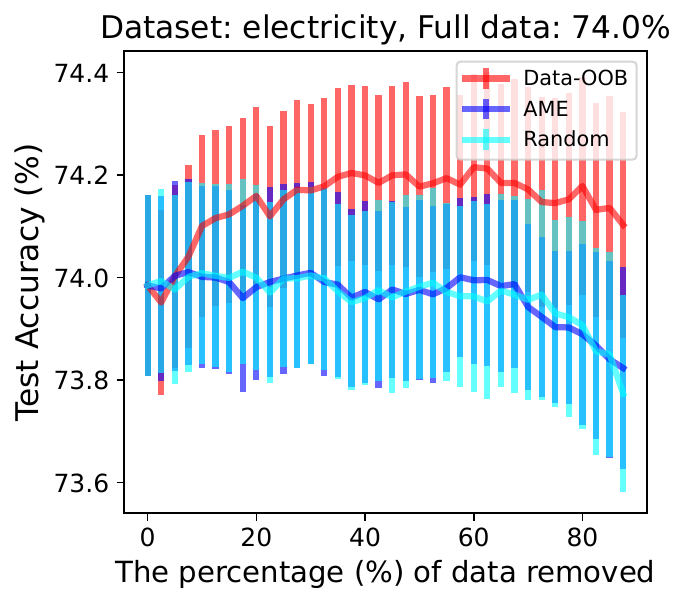}
        \includegraphics[width=0.225\textwidth]{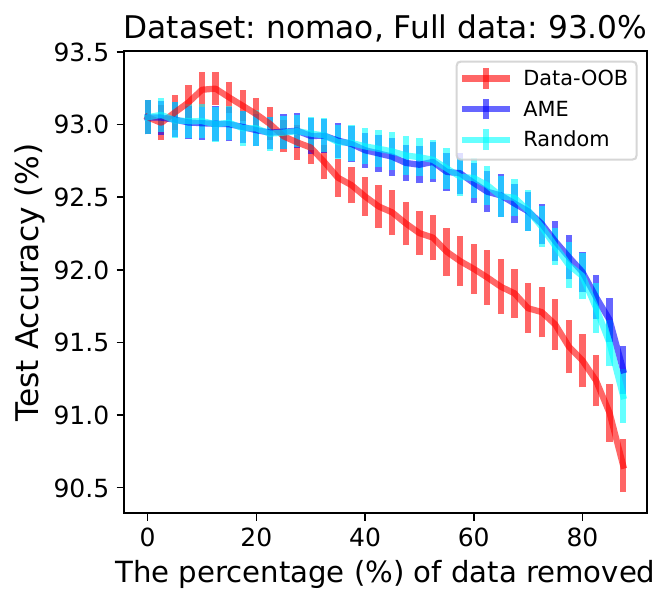}\\
        \includegraphics[width=0.225\textwidth]{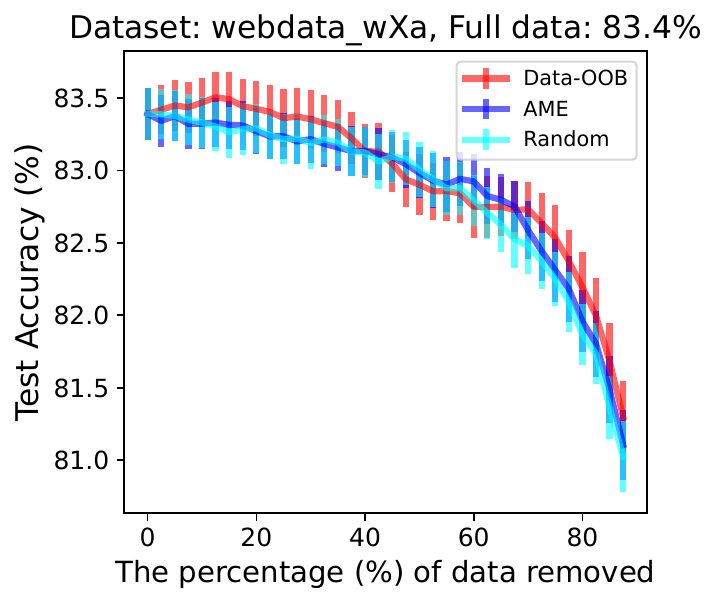}
        \includegraphics[width=0.225\textwidth]{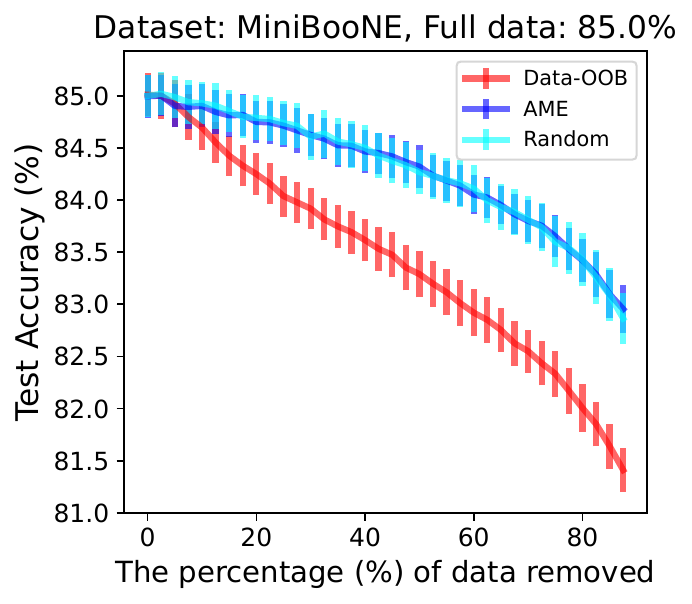}
        \includegraphics[width=0.225\textwidth]{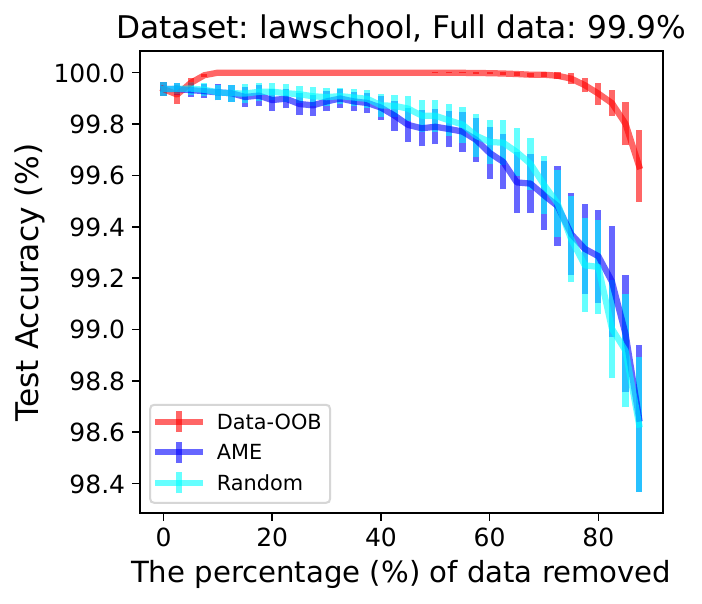}
        \includegraphics[width=0.225\textwidth]{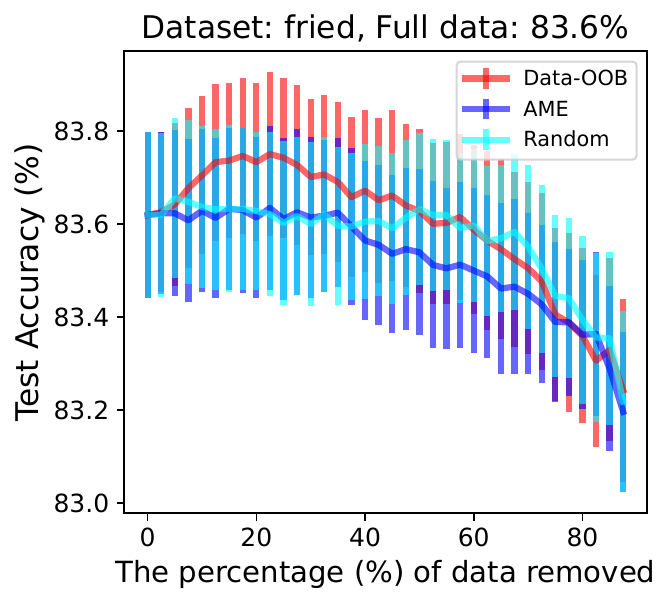}
        \caption{Test accuracy curves as a function of the percentage of data removed. We consider the four datasets when $n=10000$.}
        \label{fig:point_removal_experiment_additional_experiment10000}
    \end{subfigure}
\end{figure*}

Figures~\ref{fig:point_removal_experiment_additional_experiment1000} and~\ref{fig:point_removal_experiment_additional_experiment10000} show that point removal experiment results for the eight datasets when $n\in\{1000,10000\}$. Like Section~\ref{sec:point_removal_experiment}, we remove data points one by one starting from the lowest to the largest. Higher curves indicate better performance for data valuation. The error bar indicates a 95\% confidence interval based on 50 independent experiments. The proposed method shows a better or comparable performance than existing state-of-the-art data valuation methods in most settings. 

\newpage
\subsection{Distribution after data removal}
\label{app:distribution_after_data_removal}
\begin{figure*}[h]
    \centering
    \includegraphics[width=\textwidth]{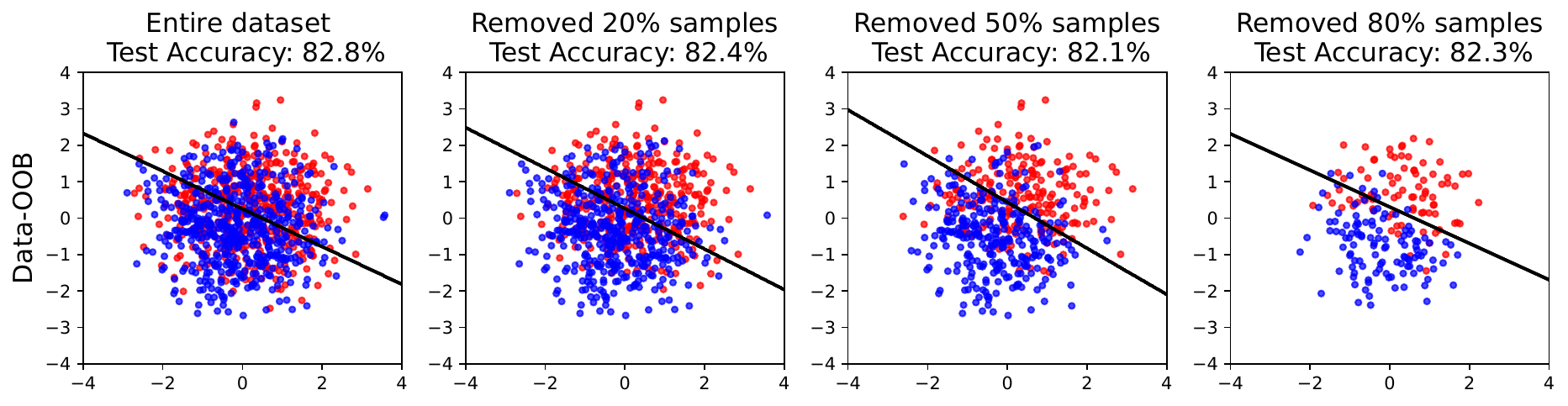}
    \includegraphics[width=\textwidth]{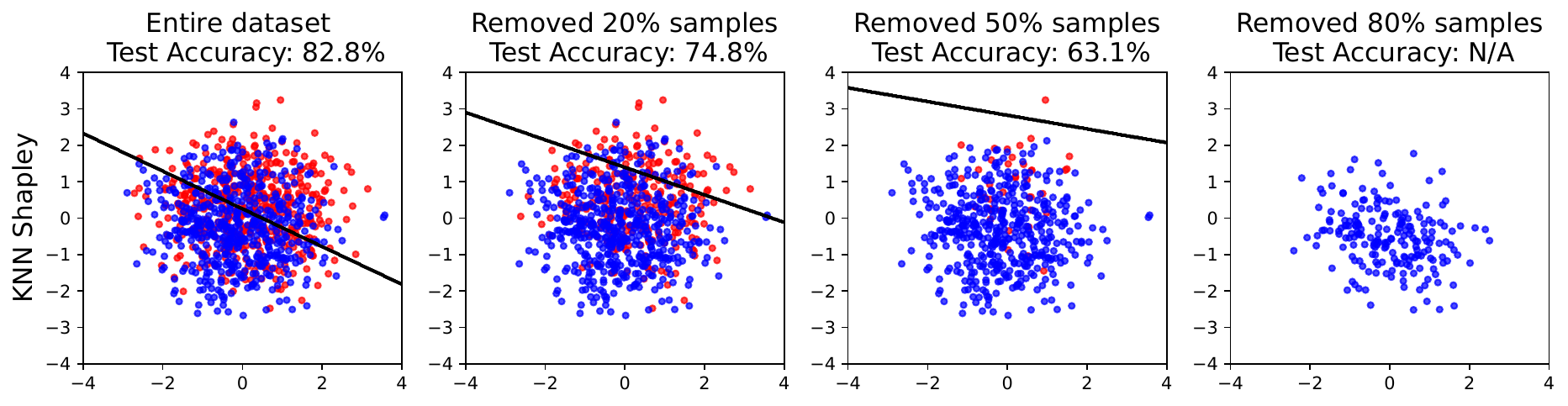}
    \includegraphics[width=\textwidth]{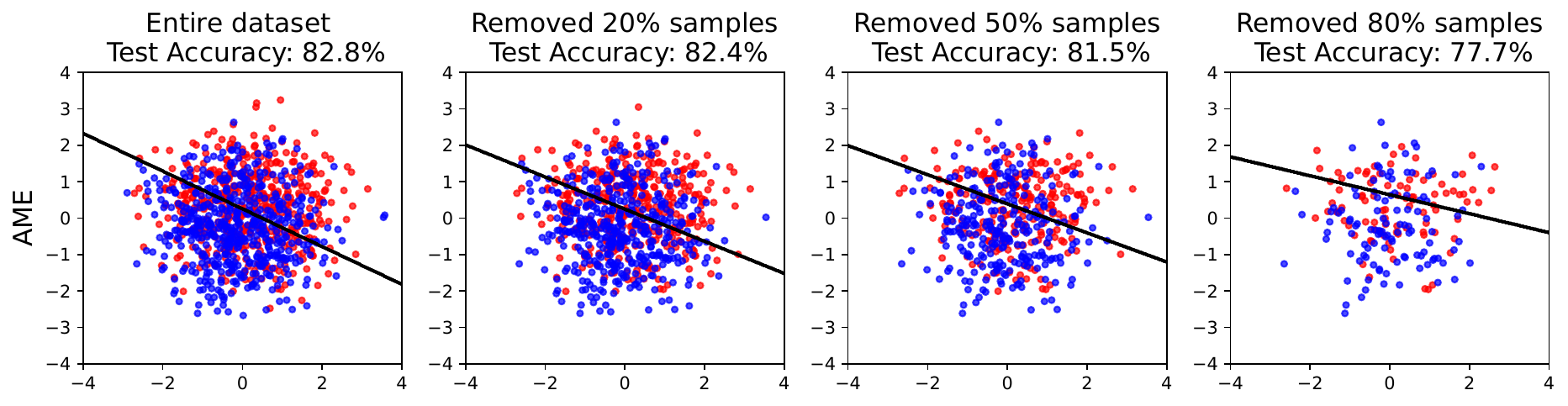} 
    \caption{Distribution after data removal for (top) \texttt{Data-OOB}, (middle) \texttt{KNN Shapley}, and (bottom) \texttt{AME}. We use the `fried' dataset. The details are given in Figure~\ref{fig:distribution_of_subsets}.}
    \label{fig:distribution_of_subsets_fried}
\end{figure*}

\begin{figure*}[h]
    \centering
    \includegraphics[width=\textwidth]{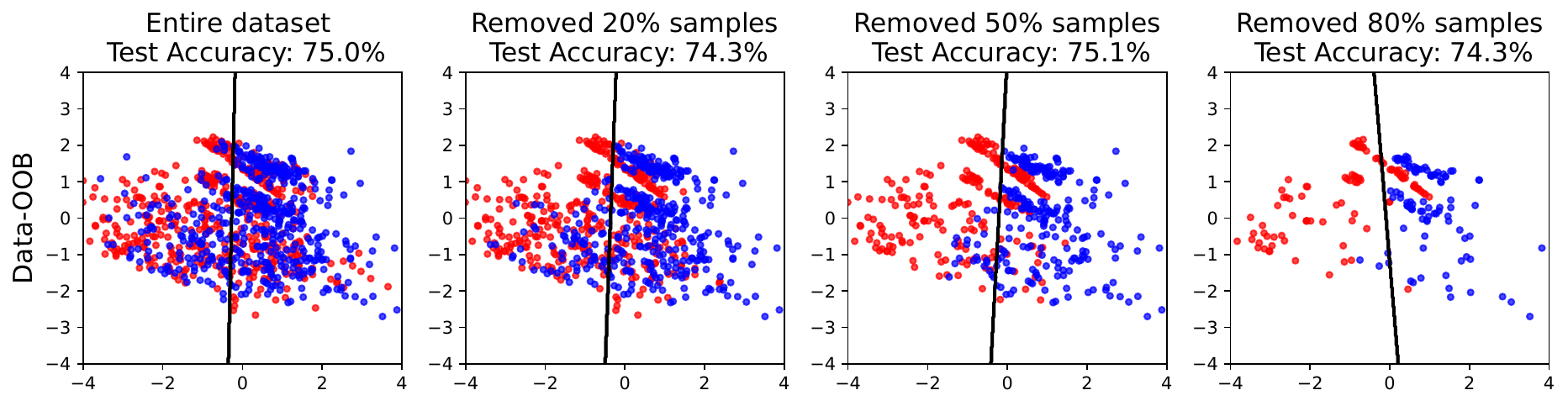}
    \includegraphics[width=\textwidth]{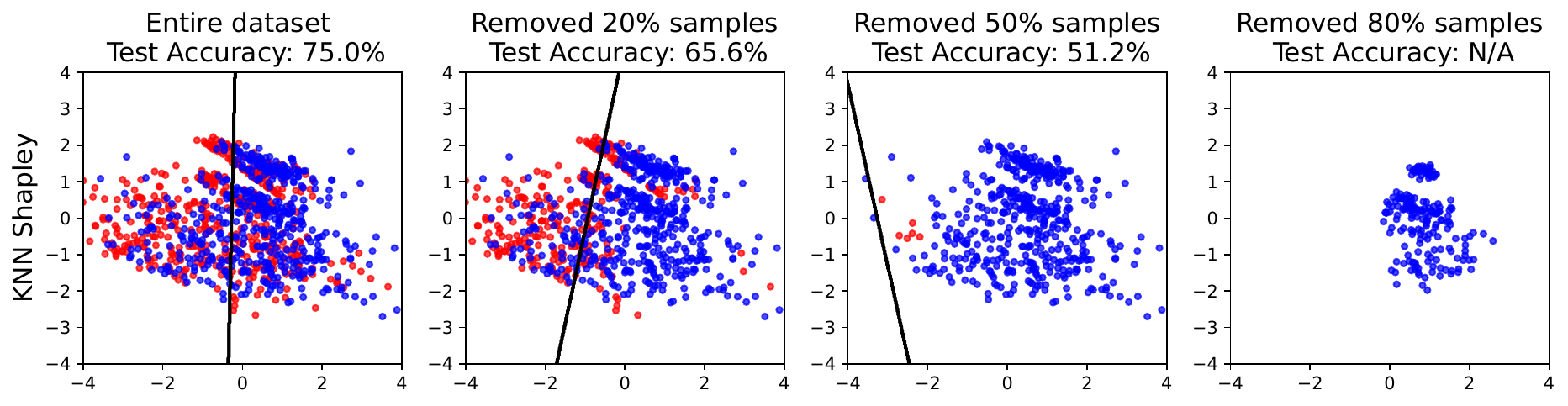}
    \includegraphics[width=\textwidth]{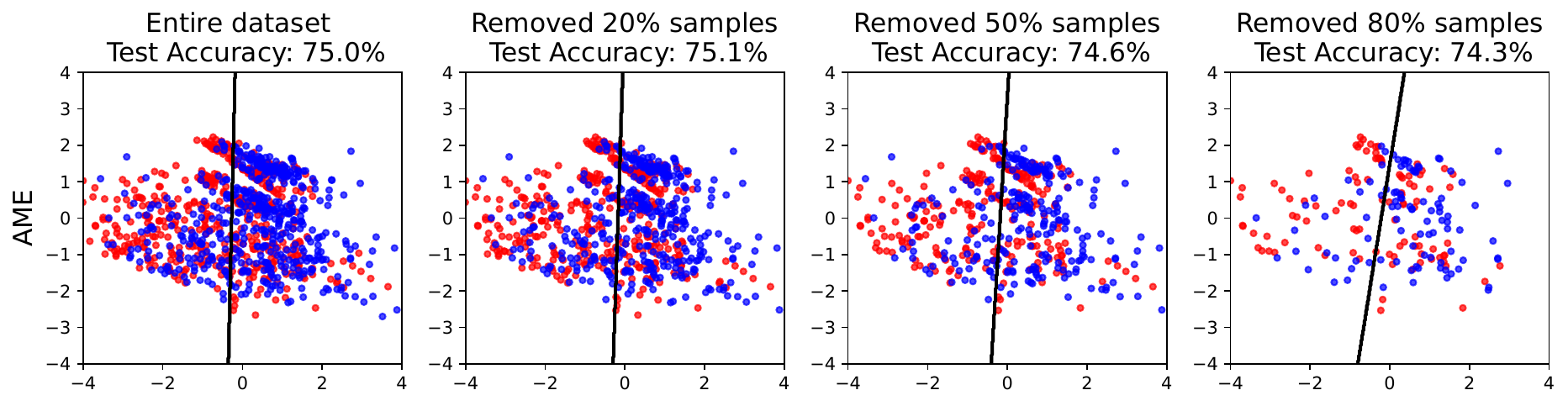} 
    \caption{Distribution after data removal for (top) \texttt{Data-OOB}, (middle) \texttt{KNN Shapley}, and (bottom) \texttt{AME}. We use the `electricity' dataset. The details are given in Figure~\ref{fig:distribution_of_subsets}.}
    \label{fig:distribution_of_subsets_electricity}
\end{figure*}

In this section, we present additional qualitative examples using the dataset `fried' in Figure~\ref{fig:distribution_of_subsets_fried} and `electricity' in Figure~\ref{fig:distribution_of_subsets_electricity}. As in Figure~\ref{fig:distribution_of_subsets}, the additional results demonstrate that \texttt{Data-OOB} is more effective in finding beneficial data points and shows competitive test accuracy even after 80\% of data removal. In contrast to \texttt{AME} that almost randomly selects data, \texttt{Data-OOB} provides a set of data points that are more sensible in that the remaining data construct clear classification regions.

\newpage

\newpage
\subsection{Robustness of the number of weak learners}
\label{app:robustness}
In this section, we show that our experimental results are robust against the number of weak learners $B$. Our results in this section show that the proposed method can achieve a similar performance with a smaller number $B$.

\paragraph{Mislabeled data detection} We compare the mislabeled data detection ability across different number of weak learners $B\in\{400, 800, 3200\}$. Table~\ref{tab:f1_score_for_different_trees} shows the F1-score of \texttt{Data-OOB} with $B \in \{400, 800, 3200\}$. In almost every dataset, the F1-score is likely to be bigger as $B$ increases, but the difference between the best and the worst is not significant. 

\begin{table}[h]
    \centering
    \begin{tabular}{l|cccccccc}
        \toprule
         Dataset & $B=400$ & $B=800$ (used in Section~\ref{sec:experiment}) & $B=3200$\\
         \midrule
         pol & $0.8806 \pm 0.0009$ & $0.8822 \pm 0.0009$ & $0.8859 \pm 0.0012$ \\ 
         jannis & $0.3351 \pm 0.0009$ & $0.3358 \pm 0.0007$ & $0.3377 \pm 0.0009$ \\ 
         lawschool & $0.9573 \pm 0.0006$ & $0.9595 \pm 0.0006$ & $0.9598 \pm 0.0005$ \\ 
         fried & $0.5398 \pm 0.0014$ & $0.5413 \pm 0.0013$ & $0.5421 \pm 0.0014$ \\ 
         electricity & $0.4348 \pm 0.0011$ & $0.4354 \pm 0.0012$ & $0.4375 \pm 0.0013$ \\ 
         2dplanes & $0.6162 \pm 0.0013$ & $0.6180 \pm 0.0013$ & $0.6188 \pm 0.0010$ \\ 
         creditcard & $0.4391 \pm 0.0010$ & $0.4420 \pm 0.0010$ & $0.4431 \pm 0.0010$ \\ 
         covertype & $0.4720 \pm 0.0011$ & $0.4790 \pm 0.0012$ & $0.4781 \pm 0.0010$ \\ 
         nomao & $0.7546 \pm 0.0014$ & $0.7564 \pm 0.0014$ & $0.7578 \pm 0.0012$ \\ 
         webdata\_wXa & $0.4151 \pm 0.0009$ & $0.4160 \pm 0.0010$ & $0.4170 \pm 0.0010$ \\ 
        \bottomrule
    \end{tabular}
    \caption{F1-score of \texttt{Data-OOB} when the number of weak learners $B \in \{400, 800, 3200\}$. The case $B=800$ is exactly the same with the one presented in the manuscript. The difference between the best and the worst is not significant, showing the robustness of the choice of $B$.}
    \label{tab:f1_score_for_different_trees}
\end{table}

\section{Proof}
\label{sec:proof}
We provide a proof for Proposition~\ref{prop:consistent_ordering}.
\begin{proof}
By Lemma 1 of \citet{efron1995cross}, the $\psi_{\mathrm{IJ}}(x_i,y_i)$ is given as
\begin{align*}
    \psi_{\mathrm{IJ}}(x_i,y_i) = \left( 2+\frac{1}{n-1} \right) \frac{\psi ((x_i, y_i), \Theta_B)  - h(\hat{\mathbb{P}})}{n} + \left( 1-\frac{1}{n} \right)^{-n} \frac{1}{B}\sum_{b=1} ^B (w_{bi}-1) q_b,
\end{align*}
where $q_b = \frac{1}{n}\sum_{j=1} ^n \mathds{1}(w_{bj}=0) T(y_j, \hat{f}_b (x_j))$ for $b \in [B]$.
Therefore, for $i \neq j$, we have
\begin{align}
    &\psi_{\mathrm{IJ}}(x_i,y_i) -  \psi_{\mathrm{IJ}}(x_j,y_j) \notag \\
    &= \left( 2+\frac{1}{n-1} \right) \frac{\psi ((x_i, y_i), \Theta_B)  - \psi ((x_j, y_j), \Theta_B) }{n} + \left( 1-\frac{1}{n} \right)^{-n} \frac{1}{B}\sum_{b=1} ^B (w_{bi}-w_{bj}) q_b.
    \label{eqn:diff_IF}
\end{align}
Let $\bar{q} = B^{-1} \sum_{b=1} ^B q_b$. Then, $\frac{1}{B}\sum_{b=1} ^B (w_{bi}-w_{bj}) q_b = \frac{1}{B}\sum_{b=1} ^B (w_{bi}-w_{bj}) (q_b - \bar{q})$ and
\begin{align*}
    \left| \frac{1}{B}\sum_{b=1} ^B (w_{bi}-w_{bj}) (q_b - \bar{q}) \right| &\leq \left( \frac{1}{B}\sum_{b=1} ^B (w_{bi}-w_{bj})^2 \right) ^{1/2} \left( \frac{1}{B}\sum_{b=1} ^B (q_b - \bar{q})^2 \right) ^{1/2}\\
    &= \left( 2-\frac{4}{n} \right)^{1/2} \{\mathrm{Var}_* (q_b)\}^{1/2}.
\end{align*}
The first inequality is from Cauchy-Schwarz inequality. The last equation is because $(w_{bi}, w_{bj})$ follows the multinomial distribution with the parameters $(n, (n^{-1}, n^{-1}))$. By the assumption on the influence function, we have
\begin{align}
    \psi_{\mathrm{IJ}}(x_i,y_i) &> \psi_{\mathrm{IJ}}(x_j,y_j) + 4 \sqrt{2} \{\mathrm{Var}_* (q_b)\}^{1/2} \notag \\
    &> \psi_{\mathrm{IJ}}(x_j,y_j) +  \left( 1-\frac{1}{n} \right)^{-n} \left( 2-\frac{4}{n} \right)^{1/2} \{\mathrm{Var}_* (q_b)\}^{1/2}.
    \label{eqn:diff_IF_2}
\end{align}
The second inequality is because $\left( 1-\frac{1}{n} \right)^{-n} \leq 4$ and $\left( 2-\frac{4}{n} \right)^{1/2} < \sqrt{2}$. Note that $\left( 1-\frac{1}{n} \right)^{-n}$ is decreasing with respect to $n \geq 2$. Therefore, Equations~\eqref{eqn:diff_IF} and~\eqref{eqn:diff_IF_2} imply that 
\begin{align*}
     \psi ((x_i, y_i), \Theta_B)  > \psi ((x_j, y_j), \Theta_B).
\end{align*}
\end{proof}

\end{document}